\pdfoutput=1

\documentclass[sigconf]{acmart}
\usepackage[utf8]{inputenc} 
\usepackage[T1]{fontenc}    

\usepackage{subfiles}
\usepackage{graphicx}
\usepackage{subcaption}
\usepackage{epstopdf}
\usepackage{marginnote}
\usepackage{enumerate}
\usepackage{color}
\usepackage{multirow}
\usepackage{xspace}
\usepackage{bm}
\usepackage{amsfonts}
\usepackage{amssymb}
\usepackage[bottom]{footmisc}

\usepackage{amsmath}
\usepackage{graphicx}
\usepackage{sidecap}
\usepackage{booktabs} 
\usepackage{colortbl} 
\usepackage{xcolor}
\usepackage{arydshln}
\usepackage{url}
\usepackage{footmisc}
\usepackage{threeparttable}

\usepackage{amsthm}
\theoremstyle{definition}
\newtheorem{definition}{Definition}

\newcommand{\V}[1]{\mathbf{#1}}

\newcommand{\bfx}{\V{x}}
\newcommand{\bfz}{\V{z}}

\newcommand{\bfZ}{\bm{\phi}}
\newcommand{\xad}{\tilde{\bfx}}
\DeclareMathOperator*{\argmax}{argmax}

\definecolor{lightgray}{gray}{0.95}

\newcommand{\vikash}[1]{\textcolor{red}{(Vikash: #1)}}

\newcommand\blfootnote[1]{%
  \begingroup
  \renewcommand\thefootnote{}\footnote{#1}%
  \addtocounter{footnote}{-1}%
  \endgroup
}

\renewcommand\footnotetextcopyrightpermission[1]{} 
\setcopyright{none}
\acmConference[]{}{}{}
\acmYear{}
\acmDOI{}

\newcolumntype{C}[1]{>{\centering\let\newline\\\arraybackslash\hspace{0pt}}m{#1}}
\newcolumntype{L}[1]{>{\raggedright \let\newline\\\arraybackslash\hspace{0pt}}m{#1}}

\begin{document}
\newcommand{\oodAdvExamples}[0]{OOD adversarial examples\xspace}    
\newcommand{\oodattack}[0]{OOD attack\xspace} 
\newcommand{\oodattacks}[0]{OOD attacks\xspace}
\newcommand{\oodlong}[0]{out-of-distribution\xspace}
\newcommand{\ood}[0]{OOD\xspace}
\newcommand{\intphoto}[0]{Internet Photographs\xspace}

\newcommand{\pgdx}[0]{\textsf{PGD-xent}}
\newcommand{\pgdcw}[0]{\textsf{PGD-CW}}

\newcommand{\bcomment}[1]{}

\newcommand{\mcX}{\mathcal{X}}
\newcommand{\mcY}{\mathcal{Y}}
\newcommand{\R}{\mathbb{R}}



\begin{minipage}{\textwidth}
\vspace{-100pt}
\centerline{\Huge \bf Better the Devil you Know: An Analysis of Evasion Attacks using }
\centerline{\Huge \bf Out-of-Distribution Adversarial Examples}
\medskip
\centerline{\Large Vikash Sehwag$^{1*}$, Arjun Nitin Bhagoji$^{1*}$, Liwei Song$^{1*}$, Chawin Sitawarin$^{2}$}
\centerline{\Large Daniel Cullina$^{1}$, Mung Chiang$^{3}$, Prateek Mittal$^{1}$} 
\medskip
\centerline{ \it $^{1}$Princeton University, $^{2}$UC Berkeley, $^{3}$Purdue University} 
\end{minipage}
\title{\vspace{10pt}}


\vspace*{-50pt}
\begin{abstract}
A large body of recent work has investigated the phenomenon of evasion attacks using adversarial examples for deep learning systems, where the addition of norm-bounded perturbations to the test inputs leads to incorrect output classification. Previous work has investigated this phenomenon in closed-world systems where training and test inputs follow a pre-specified distribution. However, real-world implementations of deep learning applications, such as autonomous driving and content classification are likely to operate in the open-world environment. In this paper, we demonstrate the success of open-world evasion attacks, where adversarial examples are generated from out-of-distribution inputs (\oodAdvExamples). In our study, we use 11 state-of-the-art neural network models trained on 3 image datasets of varying complexity. We first demonstrate that state-of-the-art detectors for out-of-distribution data are not robust against OOD adversarial examples. We then consider 5 known defenses for adversarial examples, including state-of-the-art robust training methods, and show that against these defenses, \ood adversarial examples can achieve up to 4$\times$ higher target success rates compared to adversarial examples generated from in-distribution data. We also take a quantitative look at how open-world evasion attacks may affect real-world systems. Finally, we present the first steps towards a robust open-world machine learning system. \end{abstract}

{\let\newpage\relax\maketitle}



\blfootnote{$^{*}$Equal contribution. Corresponding author (vvikash@princeton.edu)}

\section{Introduction}

\begin{table*}[!htb]
\caption{\textbf{Distinguishing between the key properties of unmodified and adversarial in-distribution and \oodlong (OOD) images for CIFAR-10 trained models}. In an open-world environment, the adversary can construct adversarial examples using OOD inputs, called \oodAdvExamples. In contrast to non-modified OOD inputs, \oodAdvExamples achieve high confidence targeted classification. Mitigation strategies such as robust training and OOD detection are largely ineffective against \oodAdvExamples.}

\label{tab: visual_demo}
\resizebox{\linewidth}{!}{%
\begin{tabular}{L{6cm}ccccc}
\toprule
\begin{tabular}[c]{@{}c@{}}\hspace{50 pt}Input Images\end{tabular} & Category & \begin{tabular}[c]{@{}c@{}}Predicted\\ class\end{tabular} & \begin{tabular}[c]{@{}c@{}}Mean output \\ Classification confidence\end{tabular} & \begin{tabular}[c]{@{}c@{}}Mitigation using\\ robust training\end{tabular} & \begin{tabular}[c]{@{}c@{}}Mitigation using\\ OOD detectors\end{tabular} \\
\midrule
\parbox[l]{6mm}{\includegraphics[width=2.1in]{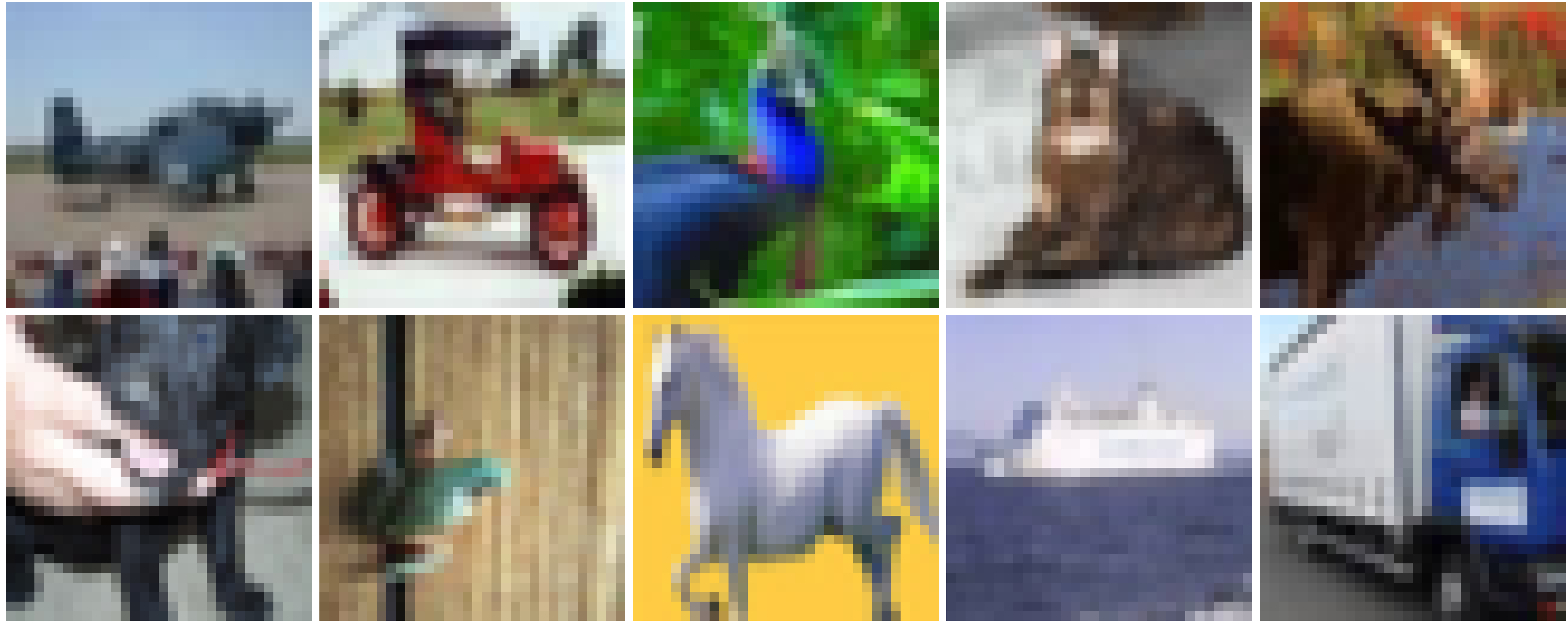}} & \begin{tabular}[c]{@{}c@{}} In-distribution \\ (unmodified)\end{tabular}& \begin{tabular}[c]{@{}c@{}}Airplane, Automobile, Bird, Cat, Deer,\\ Dog, Frog, Horse, Ship, Truck\end{tabular}  & 0.99 & N.A. & N.A.\\ \midrule
\parbox[c]{6em}{\includegraphics[width=2.1in]{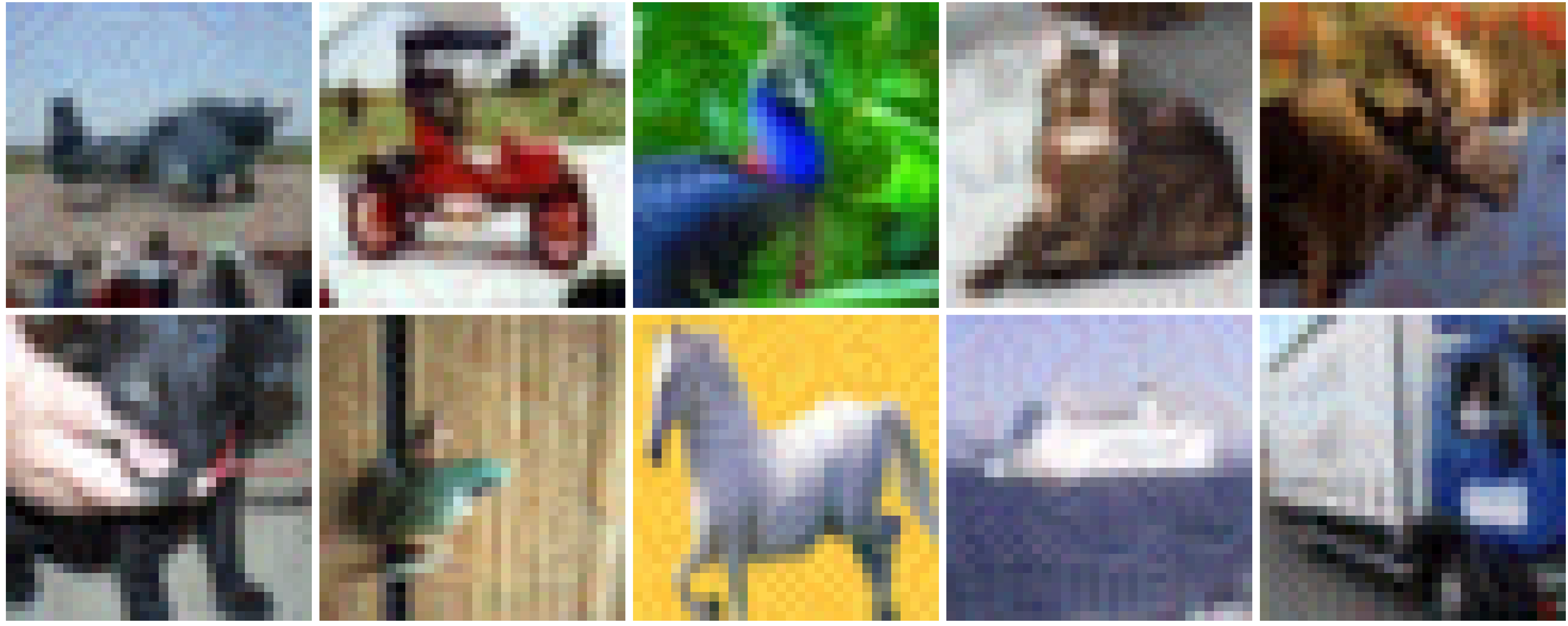}} & \begin{tabular}[c]{@{}c@{}} In-distribution \\ (adversarial)\end{tabular} & Cat & 1.00 & \checkmark & N.A.\\ \midrule
\parbox[c]{6em}{\includegraphics[width=2.1in]{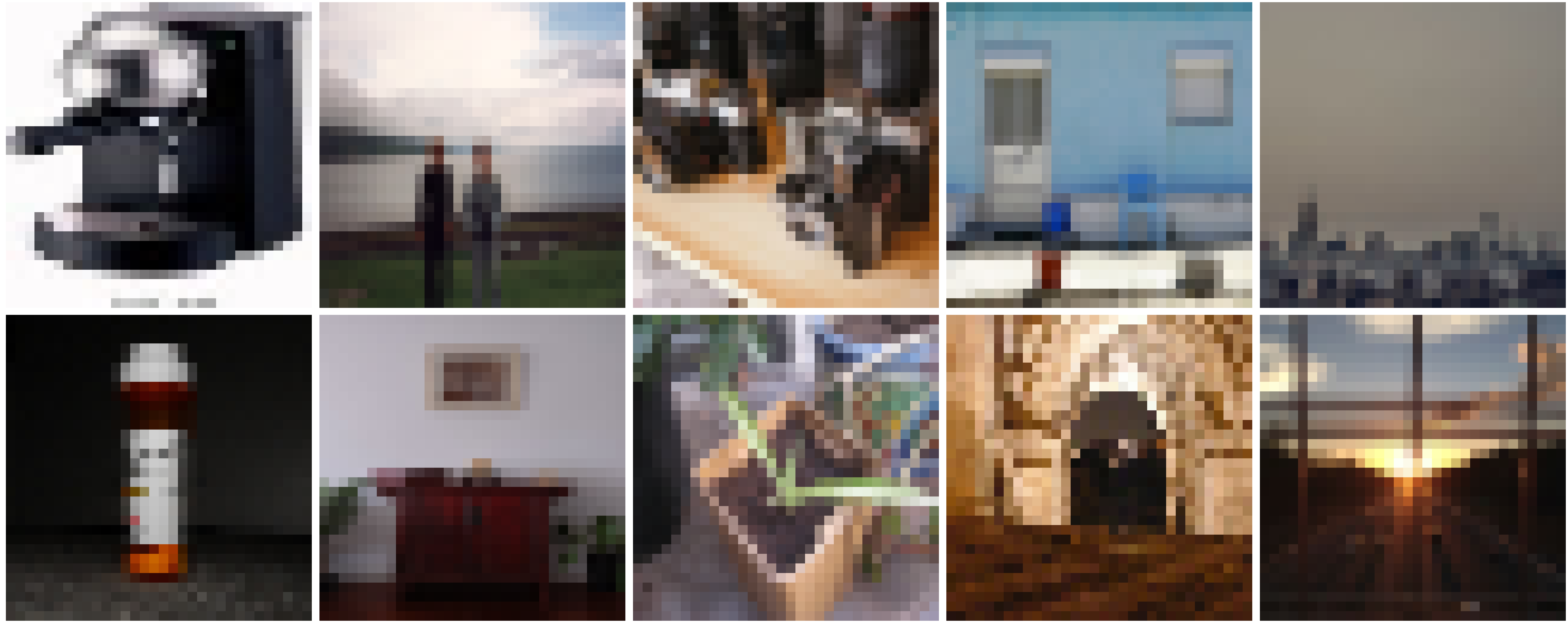}} & \begin{tabular}[c]{@{}c@{}} Out-distribution \\ (unmodified)\end{tabular} & \begin{tabular}[c]{@{}c@{}}Truck, Horse, Truck, Truck, Ship, \\ Ship, Truck, Frog, Bird, Airplane\end{tabular} & 0.53 & \text{\sffamily X} & \checkmark\\ \midrule
\parbox[c]{6em}{\includegraphics[width=2.1in]{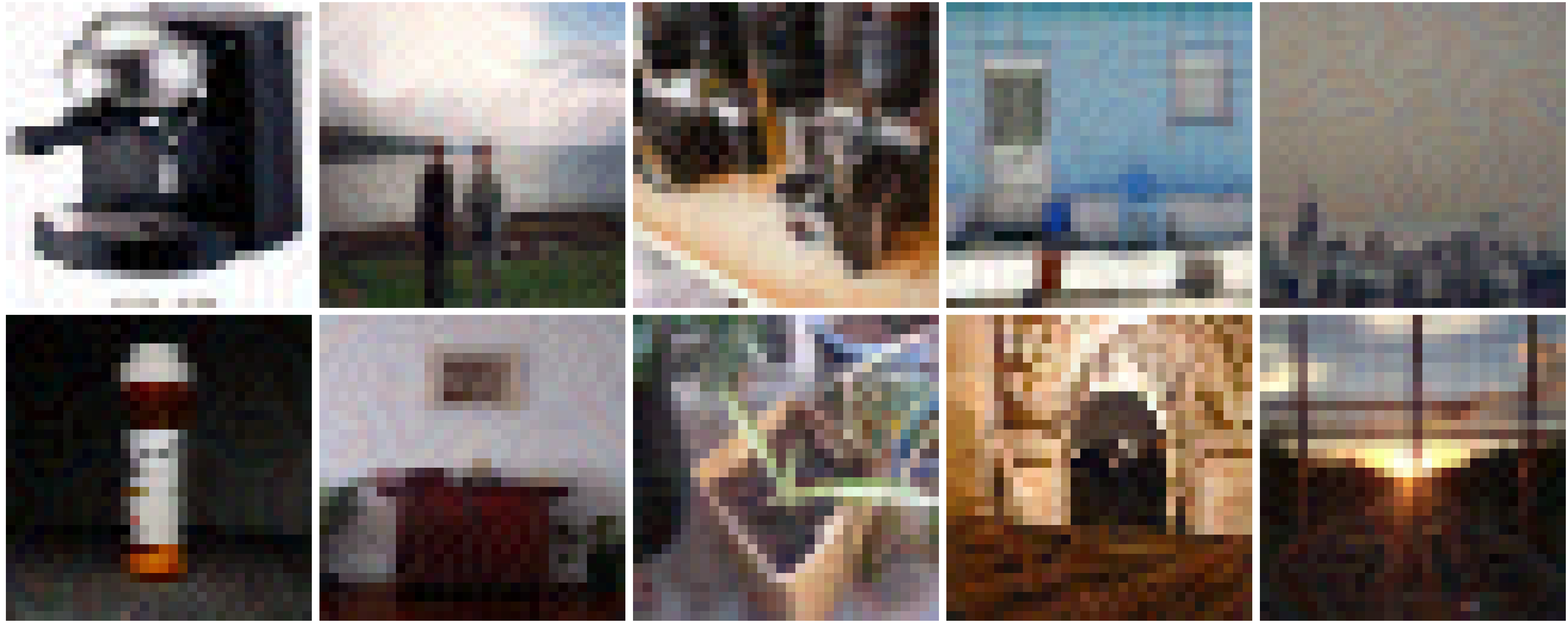}} & \begin{tabular}[c]{@{}c@{}} Out-distribution \\ (adversarial)\end{tabular}& Cat & 1.00 & \text{\sffamily X} & \text{\sffamily X} \\ \bottomrule
\end{tabular}
}
\vspace{-10pt}
\end{table*}

Machine learning (ML), spurred by the advent of deep neural networks, has become ubiquitous due to its impressive performance in domains as varied as image recognition \cite{krizhevsky2012imagenet,simonyan2014vgg}, natural language and speech processing \cite{collobert2011natural,hinton2012deep,deng2013new}, game-playing \cite{silver2017mastering,brown2017superhuman,moravvcik2017deepstack} and aircraft collision avoidance \cite{julian2016policy}. However, its ubiquity provides adversaries with both opportunities and incentives to develop strategic approaches to fool machine learning systems during both training (poisoning attacks) \cite{biggio2012poisoning,rubinstein2009stealthy,mozaffari2015systematic,jagielski2018manipulating} and test (evasion attacks) \cite{szegedy2014intriguing,goodfellow2014explaining,papernot2016limitations,moosavi2015deepfool,moosavi2016universal,carlini2017towards} phases. This paper's focus is evasion attacks which have been proposed against supervised ML algorithms used for image classification \cite{biggio2013evasion,szegedy2014intriguing,goodfellow2014explaining,carlini2017towards,papernot2016limitations,chen2017ead}, object detection \cite{xie2017adversarial,lu2017adversarial,chen2018robust}, image segmentation \cite{fischer2017adversarial,arnab_cvpr_2018}, speech recognition \cite{carlini2018audio,yuan2018commandersong} as well as other tasks \cite{cisse2017houdini,kantchelian2016evasion,grosse2017adversarial,xu2016automatically}; generative models for image data \cite{kos2017adversarial}, and even reinforcement learning algorithms \cite{kos2017delving,huang2017adversarial}. Successful instantiations of these attacks have been demonstrated in black-box \cite{szegedy2014intriguing,papernot2016transferability,papernot2016practical,liu2016delving,brendel2017decision,bhagoji_eccv_2018,chen2017zoo} as well as physical settings \cite{sharif2016accessorize,kurakin2016adversarial,Evtimov17,sitawarin2018rogue}. The vast majority of these attacks consider adversarial examples generated from inputs taken from the training or test data (referred to as in-distribution examples henceforth).

In this paper, we consider the threat posed by evasion attacks in the \emph{open-world learning model}, where the ML system is expected to handle inputs that come from the same input space as the in-distribution data, but may arise from completely different distributions, referred to as \emph{out-of-distribution} (OOD) data. Motivated by the fact that real-world applications of machine learning are likely to operate in the open-world environment, this problem has been studied widely under the context of \oodlong detection, anomaly detection, and selective prediction \cite{hendrycks2017iclr, liang2017ODIN, lee2017training, devries2018learning, jiang2018trust, lakshminarayanan2017simple, lee2018unifieddetector, zong2018deepmm, liu2018openPAC, ruff2018deepsvm, chalapathy2018OCNN, hendrycks2018outlierexpose, NIPS2018lossframework, dhamija2018agnostophobia, cvpr16_open_set_networks, bendale2015openworld, yoshihashi2018advanceOS, gunther2017opensetface, el2010backclass, chang1994backclass, geifman2017backclass}. Indeed, there have been many recent attempts extending the state-of-the-art deep learning systems to the open-world environment ~\cite{cvpr16_open_set_networks, bendale2015openworld, yoshihashi2018advanceOS, gunther2017opensetface}. However, we note that work on the open-world learning model has not considered the presence of adversaries, while work on adversarial examples has been restricted to generating them from in-distribution adversarial examples. We close this gap by considering \emph{adversaries operating in the open-world learning setting} that carry out evasion attacks by modifying OOD data (compliant with the environment/application constraints, if any) with the aim of inducing high-confidence targeted misclassification. 

To carry out evasion attacks on open-world learning models, we introduce out-of-distribution (\ood) adversarial examples: \textit{adversarial examples created by perturbing an OOD input to be classified as a target class $T$, where an OOD input is an arbitrary input drawn from a distribution different from training/test distribution}. We demonstrate that \oodAdvExamples are able to bypass current OOD detectors as well as defenses against adversarial examples (Table \ref{tab: visual_demo}). Under the taxonomy of attacks on ML systems laid out by Huang et al. \cite{huang2011adversarial}, \ood adversarial examples are a form of \emph{exploratory integrity attacks} with the intent of \emph{targeted} misclassification of input data. Intuitively, we would expect a well-behaved classifier to classify an \ood adversarial example with low confidence, since it has never encountered an example from that portion of the input space before. This intuition underlies the design of state-of-the-art OOD detectors. However, \oodAdvExamples constitute an integrity violation for classifiers as well as OOD detectors as they induce high confidence targeted misclassification in state-of-the-art classifiers, which is unwanted system behavior. While previous work \cite{goodfellow2014explaining,nguyen2015deep,sharif2016accessorize,sitawarin2018rogue} has hinted at the possibility of generating adversarial examples without the use of in-distribution data, we are the first to rigorously examine \ood adversarial examples and their impact on ML classifiers, including those secured with state-of-the-art defenses, as well as OOD detectors. 

\subsection{Contributions}
We now detail our contributions in this paper.

\noindent \textbf{Introduction and analysis of open-world evasion attacks:} We introduce open-world evasion attacks and propose \oodAdvExamples to carry out these attacks. We study their impact on state-of-the-art classifiers, both defended and undefended, as well as on \ood detectors. We consider 11 different neural networks trained across 3 benchmark image datasets of varying complexity. For each dataset used for training, we use 5 \ood datasets for the generation of \ood adversarial examples in order to determine the effect of the choice of \ood data on robustness.

\noindent \textbf{Evading state-of-the-art \ood detectors:} We evaluate the robustness of the state-of-the-art OOD detection mechanisms ODIN \cite{liang2017ODIN} and Confidence-calibrated classifiers \cite{lee2017training} against \oodAdvExamples. Our results show that the OOD detectors, which can detect close to 85$\%$ of non-modified OOD inputs, fail to detect a significant percentage of \oodAdvExamples~(up to 99.8$\%$).

\noindent \textbf{Bypassing state-of-the-art defenses for in-distribution attacks:} Although state-of-the-art defenses such as iterative adversarial training \cite{madry_towards_2017} and provably robust training with the convex outer polytope \cite{kolter2017provable} are promising approaches for the mitigation of in-distribution attacks, their performance significantly degrades with the use of OOD adversarial examples. We demonstrate that \oodAdvExamples can achieve a significantly higher target success rate (up to 4$\times$ greater) than that of adversarial examples generated from in-distribution data. Further, we demonstrate that \oodAdvExamples are able to evade adversarial example detectors such as feature squeezing \cite{xu2017feature} and MagNet \cite{meng2017magnet}, with close to 100$\%$ success rate (similar to in-distribution adversarial examples). We also show this for the Adversarial Logit Pairing defense \cite{kannan2018adversarial}.

\noindent \textbf{\ood adversarial examples in the real world:} We demonstrate the success of OOD adversarial examples in real-world settings by targeting a content moderation service provided by Clarifai \cite{clarifai}. We also show how physical \ood adversarial examples can be used to fool traffic sign classification systems. 

\noindent \textbf{Towards robust open-world learning:} We explore if it is possible to increase the robustness of defended models by including a small number of OOD adversarial examples during robust training. Our results show that such an increase in robustness, even against OOD datasets excluded in training, is possible. 

Overall, we demonstrate that considering the threats posed to open-world learning models is imperative for the secure deployment of ML systems in practice.
\bcomment{
\begin{table*}[!htb]
    \renewcommand{\arraystretch}{1.4}
    \caption{Summary of defenses and results on CIFAR-10 trained models. Novel results and empirical conclusions from this paper are in \textbf{bold}. The last column shows the successful \oodAdvExamples for each defense, where each of them is generated to be classified as an $airplane$ by the network. It can be observed that the added perturbations are imperceptible.
    }
    \label{tab: summary_table}
    \resizebox{\textwidth}{!}{%
        \begin{tabular}{lccccr}
            \toprule
            & & \multicolumn{3}{c}{Behavior on data type} & \\ \cline{4-5}
            Defense Type & Defense Name &  & \begin{tabular}[c]{@{}c@{}} In-distribution adversarial \\ (white-box, adaptive) \end{tabular}& \textbf{\begin{tabular}[c]{@{}c@{}} Out-of-distribution adversarial \\(white-box, adaptive) \end{tabular}} & \begin{tabular}[c]{@{}c@{}} Representative \\ \oodAdvExamples \end{tabular}\\ \midrule
            \multirow{3}{*}{\parbox[t]{2mm}{\rotatebox[origin=c]{0}{N.A.}}} & \multirow{3}{*}{Undefended} & & \multirow{3}{*}{\begin{tabular}[c]{@{}c@{}}Not robust \cite{madry_towards_2017,carlini2017towards} \\ Rate: 100.0, Conf: 1.00 \end{tabular}} & \multirow{3}{*}{\begin{tabular}[c]{@{}c@{}}Not robust\\ Rate: 100.0, Conf: 1.00 (ImageNet)\end{tabular}} & \multirow{3}{*}{\includegraphics[width=0.95in
            ]{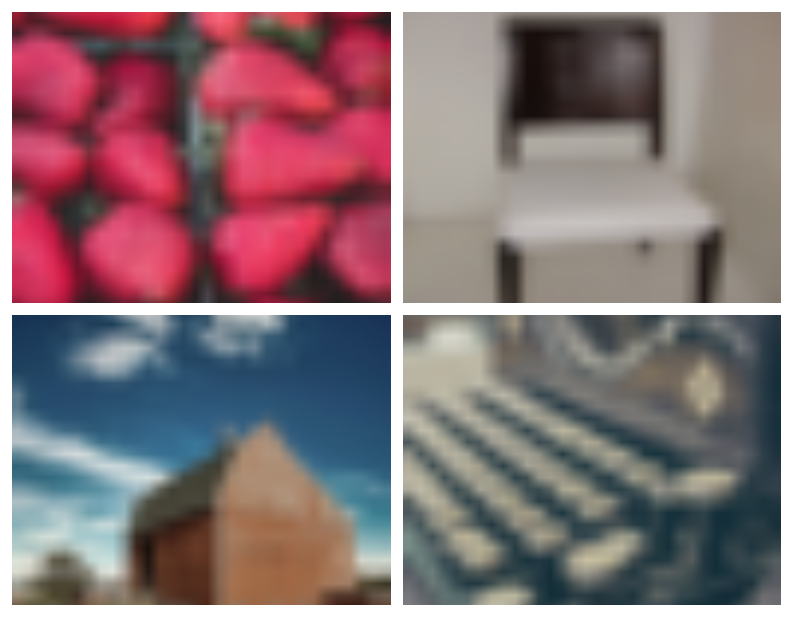}}\\
            & & & \begin{tabular}[c]{@{}c@{}} \end{tabular} & \begin{tabular}[c]{@{}c@{}} \end{tabular} & \\
            & & & \begin{tabular}[c]{@{}c@{}} \end{tabular} & \begin{tabular}[c]{@{}c@{}} \end{tabular} & \\ \midrule
            \parbox[t]{2mm}{\multirow{2}{*}{\rotatebox[origin=c]{0}{\begin{tabular}[c]{@{}c@{}} Robust\\ Training \end{tabular}}}} & Iterative adv. training \cite{madry_towards_2017} & & \begin{tabular}[c]{@{}c@{}}Somewhat robust \cite{madry_towards_2017}\\ Rate: 22.9, Conf: 0.81 \end{tabular} & \begin{tabular}[c]{@{}c@{}}Not robust \\ {Rate: 87.9, Conf: 0.86 (Gaussian Noise)} \end{tabular} & \multirow{2}{*}{\includegraphics[width=0.95in
            ]{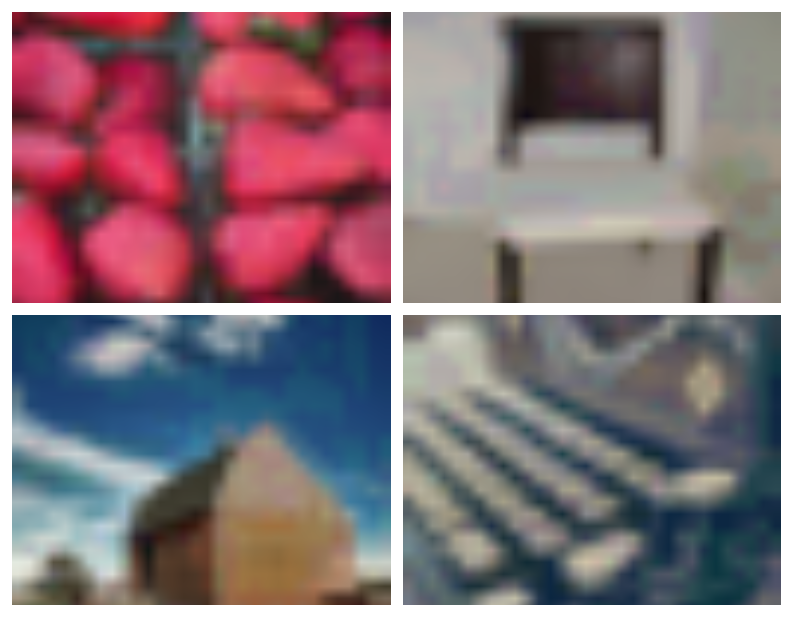}}\\ 
            & Convex polytope relaxation \cite{kolter2017provable} & & \begin{tabular}[c]{@{}c@{}}Provably robust \cite{kolter2017provable}\\ Rate: 15.1, Conf: 0.41 \end{tabular} & \begin{tabular}[c]{@{}c@{}}Somewhat robust \\ Rate: 29.1, Conf: 0.32 (Gaussian Noise)\end{tabular} & \\ \midrule
            \parbox[t]{2mm}{\multirow{2}{*}{\rotatebox[origin=c]{0}{\begin{tabular}[c]{@{}c@{}} Adversarial \\ Example\\ Detection \end{tabular}}}} & Feature Squeezing \cite{xu2017feature} & & \begin{tabular}[c]{@{}c@{}}Not robust \cite{he2017ensembleweak}\\ Rate: 100.0, Conf: 0.99 \end{tabular} & \begin{tabular}[c]{@{}c@{}}Not robust \\ Rate: 100.0, Conf: 0.99 (ImageNet)  \end{tabular} & \multirow{2}{*}{\includegraphics[width=0.95in
            ]{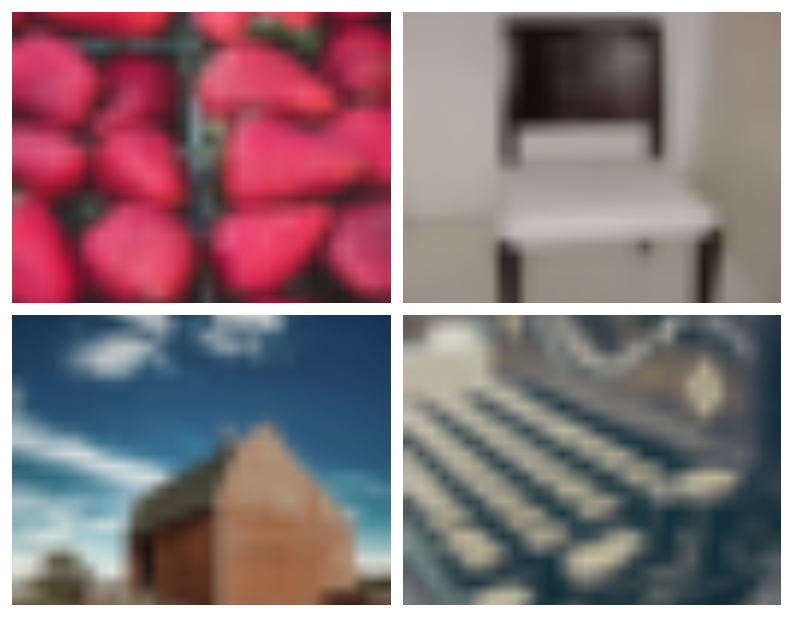}}\\
            & MagNet \cite{meng2017magnet} & & \begin{tabular}[c]{@{}c@{}}Not robust \cite{carlini2017magnet}\\ Rate: 90.1, Conf: 0.97  \end{tabular}  & \begin{tabular}[c]{@{}c@{}}Not robust \\ Rate: 97.3 , conf: 0.98 (Random Photographs)\end{tabular} & \\ \midrule
            \parbox[t]{2mm}{\multirow{2}{*}{\rotatebox[origin=c]{0}{\begin{tabular}[c]{@{}c@{}} OOD\\ Detection \end{tabular}}}} & ODIN \cite{liang2017ODIN} & & N.A. & \begin{tabular}[c]{@{}c@{}}Not robust \\ Rate: 81.6, Conf: 0.97 (Random Photographs)\end{tabular} & \multirow{2}{*}{\includegraphics[width=0.95in
            ]{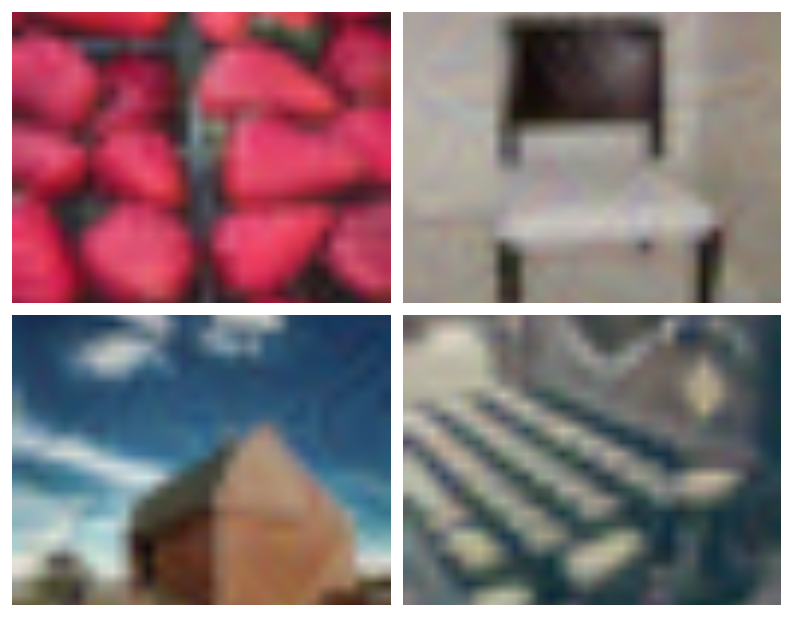}} \\
            & Confidence-calibrated \cite{lee2017training} & & N.A. & \begin{tabular}[c]{@{}c@{}} Somewhat robust \\ Rate: 47.1 , Conf: 0.99 (VOC12) \end{tabular} & \\ \bottomrule
        \end{tabular}
    }
\end{table*}
}

\section{Background and Related Work}\label{sec:background}
In this section we present the background and related work on adversarial examples generated from in-distribution data, defenses against evasion attacks and state-of-art detection methods for unmodified OOD data.

\subsection{Supervised classification}\label{subsec: back_ml}
Let $\mcX$ be a space of examples and let $\mcY$ be a finite set of classes.
A classifier is a function $f(\cdot): \mcX \rightarrow \mcY$.
Let $\mathbb{P}(\mcY)$ be the set of probability distributions over $\mcY$.
In our setting, a classifier is always derived from a function $g(\cdot): \mcX \rightarrow \mathbb{P}(\mcY)$ that provides confidence information, i.e. $f(x) = \argmax_{i \in \mcY} g(x)(i)$. In particular, for DNNs, the outputs of the penultimate layer of a neural network $f$, representing the output of the network computed sequentially over all preceding layers, are known as the logits. We represent the logits as a vector $\bfZ^f(\bfx) \in \mathbb{R}^{|\mathcal{Y}|}$. The classifier is trained by minimizing the empirical loss $\frac{1}{n}\sum_{i=1}^{n}\ell_{g}(\bfx_i,y_i)$ over $n$ samples $\{(\bfx_1,y_1), \ldots (\bfx_n,y_n)\}$ (training set), where $\ell_{g}(\cdot,\cdot)$ is a loss function such as the cross-entropy loss \cite{goodfellow2016deep} that depends on the output confidence function $g(\cdot)$ \cite{murphy2012machine}. The training set is drawn from a distribution $P^{\text{in}}_{X,Y}$ over the domain $\mcX \times \mathcal{Y}$. The marginal distribution over the space of examples $\mcX$ is represented as $P^{\text{in}}_{X}$. These samples usually represent an application-specific set of concepts that the classifier is being trained for.


\subsection{Evasion attacks}\label{subsec: back_attack}
We focus on test phase or \emph{evasion attacks}. 

\subsubsection{Adversarial examples generated from in-distribution data}
Evasion attacks have been demonstrated to be highly successful for a number of classifiers~\cite{biggio2013evasion,szegedy2014intriguing,goodfellow2014explaining,moosavi2015deepfool,papernot2016limitations,carlini2017towards, kurakin2016adversarial,Evtimov17,sitawarin2018rogue, suciu2018FAILUsenix, wang2018trasferattack}. All of these aim to modify benign, \emph{in-distribution} examples $\bfx \sim P^{\text{in}}_{X}$ by adding a imperceptible perturbation to them such that the modified examples $\xad$ are \emph{adversarial} \cite{szegedy2014intriguing}. The adversary's aim is to ensure that these adversarial examples are successfully misclassified by the ML system in a targeted class (targeted attack), or any class other than the ground truth class (untargeted attack). We focus entirely on \emph{targeted attacks} since these are more realistic from an attacker's perspective and are strictly harder to carry out than untargeted attacks. To generate a successful \emph{targeted} adversarial example $\xad$ for class $T$ starting from a benign example $\bfx$ for a classifier $f$, the following optimization problem must be solved
\begin{equation}\label{eq: comb}
f(\xad) = T \quad \text{s.t.} \quad d(\xad,\bfx) < \epsilon
\end{equation}
where $d(\cdot,\cdot)$ is an appropriate distance metric for inputs from the input domain $\mathcal{X}$ used to model imperceptibility-based adversarial constraints \cite{goodfellow2014explaining,carlini2017towards}. The distance metric imposes an $\epsilon$-ball constraint on the perturbation. The optimization problem in Eq. \ref{eq: comb} is combinatorial and thus difficult to solve. In practice, a relaxed version using an appropriate adversarial loss function $\ell^{\text{adv}}_{g}(\bfx,T)$ derived from the confidence function $g(\cdot)$ is used and solved with an iterative optimization technique \cite{szegedy2014intriguing,goodfellow2014explaining,carlini2017towards,madry_towards_2017,athalye2018obfuscated,moosavi2015deepfool}. Details of the state-of-the-art attack methods we use are in Section \ref{subsec: setup_attack}.

\subsubsection{Real-world attacks}\label{subsubsec: real-world}
Work on adversarial examples has also examined the threat they pose in real-world settings. One line of work has been to analyze attacks on black-box models hidden behind APIs \cite{bhagoji_eccv_2018,chen2017zoo,ilyas2018limitedquery,brendel2017decision} where access to the internals of a model is often unavailable due to privacy concerns. In these settings, gradient estimation techniques or other black-box optimization techniques such as particle swarm optimization \cite{bhagoji_eccv_2018} are used to generate adversarial examples. Further, the possibility of using physically realized adversarial examples \cite{Sitawarin17,Evtimov17,Athalye17} to attack ML systems has also been considered. Since the image capture process introduces artifacts related to brightness, scaling and rotation, these have to be considered in the adversarial example generation process, which is done using the expectation over transformations method \cite{Athalye17}. We examine both of these attacks for \ood adversarial examples in Section \ref{subsec: real-world}.
    
\subsubsection{Threat models}\label{subsubsec:back_threat}
Based on the degree of access the adversary has to the classifier under attack, different threat models can be defined. The two threat models we consider in this paper are:\\
\noindent \textbf{White-box:} In this threat model, the adversary has full access to the 
classifier under attack, including any possible defenses that may have been employed, as well as the training and test data \cite{szegedy2014intriguing,goodfellow2014explaining,carlini2017towards,athalye2018obfuscated}. This is the threat model under which the evaluation of defenses has been recommended in the literature \cite{carlini2017towards,athalye2018obfuscated} in order to avoid reliance on `security through obscurity' and is the \emph{primary threat model} used in this paper.\\
\noindent \textbf{Black-box with query access:} This is a more restrictive threat model, where the adversary is only assumed to have access to the output probability distribution $g(\cdot)$ for any input $\bfx \in \mathcal{X}$ \cite{dang2017extracting, narodytska2016simple, chen2017zoo, bhagoji_eccv_2018, ilyas2018limitedquery}. No access to either the model structure and parameters or the training and test data is assumed. This is a \emph{secondary threat model} we use to demonstrate attacks on cloud-based classifiers \cite{clarifai} (Section \ref{subsec: real-world}).

\subsection{Defenses against evasion attacks}\label{subsec: back_defenses}
In order to defend against evasion attacks, there are broadly two approaches used in the literature as noted in the review by Papernot et al. \cite{papernot2016towards}. The first, namely robust training, seeks to \emph{embed resilience} into the classifier during training by modifying the standard loss $\ell_{g}(\cdot,\cdot)$ used during the training process to one that accounts for the presence of an adversary at test time \cite{goodfellow2014explaining,tramer2017ensemble,madry_towards_2017,kolter2017provable,raghunathan2018certified}. The second approach adds additional pre-processing steps at the input in order to detect or defend against adversarial examples \cite{bhagoji2017dimensionality,meng2017magnet,xu2017feature, guo2017countering, xie2017mitigating, samangouei2018defense, song2017pixeldefend}. 

\subsubsection{Robust training} \label{subsubsec: back_robust_training}
Robust training can be performed either in a heuristic manner through the use of adversarial training \cite{goodfellow2014explaining,kurakin2017adversarial,tramer2017ensemble,madry_towards_2017,kannan2018adversarial} or in a provable manner with the use of neural network verification \cite{ehlers2017planet,gowal2018effectiveness, wang2018formal, wang2018efficient, wang2018mixtrain, weng2018towards, xiao2018training, gehr2018AI2}. Since exact verification of state-of-the-art neural networks is inefficient due to their size and complexity, convex relaxations are used for training provably robust networks\cite{kolter2017provable, wong2018scaling, raghunathan2018certified,sinha2018certifying}.

\noindent \textbf{Adversarial training:} 
These heuristic methods defend against adversarial examples by modifying the loss function such that it incorporates both clean and adversarial inputs. \begin{small}\begin{align}\label{eq: adv_train}
    \tilde{\ell}_{g}(\bfx,y) = \alpha \ell_{g}(\bfx,y) + (1- \alpha) \ell_{g}(\xad,y),
    \end{align}\end{small}
where $y$ is the true label of the sample $\bfx$.\par
In this paper, we consider the robustness of networks trained using \emph{iterative adversarial training} \cite{madry_towards_2017}, which uses adversarial examples generated using Projected Gradient Descent (PGD). This method has been shown to be empirically robust to adaptive white-box adversaries using adversarial examples generated from in-distribution data which use the same $L_p$ norm \cite{athalye2018obfuscated} for models trained on the MNIST \cite{lecun1998mnist} and CIFAR-10 \cite{krizhevsky2014cifar} datasets.

\noindent \textbf{Provable robustness using convex relaxations:} We focus on the approach of Kolter and Wong \cite{kolter2017provable,wong2018scaling} which scales to neural networks on the CIFAR-10 dataset \cite{krizhevsky2014cifar}. They aim to certify robustness in an $\epsilon$-ball around any point in the input space by upper bounding the adversarial loss with a convex surrogate. They find a convex outer approximation of the activations in a neural network that can be reached with a perturbation bounded within an $\epsilon$-ball and show that an efficient linear program can be used to minimize the worst case loss over this region. In Section \ref{subsec: strong_defenses}, we show reduced effectiveness of these two robust training based defenses for \ood adversarial examples. 

\subsubsection{Adversarial Example Detectors and {secondary} defenses} \label{subsubsec: back_adv_detection} 
 Seeking to exploit the difference in the properties of adversarial and unmodified inputs for in-distribution data, a large number of adversarial example detectors have been proposed \cite{meng2017magnet, xu2017feature, guo2017countering, xie2017mitigating, grosse2017statistical, samangouei2018defense, song2017pixeldefend}. We consider two of the most promising detectors, namely \emph{feature squeezing} \cite{xu2017feature} and \emph{MagNet} \cite{meng2017magnet}. Both methods use input pre-processing in order to distinguish between benign and adversarial examples, but feature squeezing performs detection solely based on classifier outputs, while MagNet can perform detection at both the input and output. Unfortunately, these methods are not robust against adaptive white-box adversaries generating adversarial examples from in-distribution data \cite{he2017ensembleweak,carlini2017magnet,athalye2018obfuscated}. In Section \ref{subsec: adv_detectors}, we show that this lack of robustness persists for OOD adversarial examples.
 
 Iterative adversarial training proposed by Madry et al. \cite{madry_towards_2017} does not converge for Imagenet-scale models \cite{madry_towards_2017}.  
 The \emph{Adversarial Logit Pairing} (ALP) \cite{kannan2018adversarial} defense claimed to provide robustness for Imagenet-scale models by adding an additional loss term proportional to the distance between $\bfZ(\bfx)$ and $\bfZ(\xad)$ during training, with $\xad$ generated using PGD. However, it was shown that simply increasing the number of PGD iterations used to generate adversarial examples from in-distribution data reduced the additional robustness to a negligible amount \cite{breakingALP2018}. In Section \ref{subsec: strong_defenses}, we show that this lack of robustness persists for OOD adversarial examples.


\subsection{Open-world Deep learning} \label{subsec: back_open_world}
The closed-world approach to deep learning, described in Section \ref{subsec: back_ml} operates using the assumption that both training and test data are drawn from the same application-specific distribution $P^{\text{in}}_{X,Y}$. However, in a real-world environment, ML systems are expected to encounter data at test time that is not drawn from $P^{\text{in}}_{X,Y}$ but belongs to the same input space, i.e. they encounter samples that are \emph{\oodlong} (\ood). This leads to the \emph{open-world learning model}. Thus, in order to extend supervised learning algorithms to the open-world learning model, it is critical to enable them to reject \oodlong inputs. The importance of this learning model is highlighted by the fact that a number of security and safety-critical applications such as biometric authentication, intrusion detection, autonomous driving, medical diagnosis are natural settings for the use of open-world machine learning \cite{gunther2017opensetface, chalapathy2019ADsurvey, ramanagopal2018oodAV, litjens2017OODmed}.\par

 \subsubsection{Out-of-Distribution data} To design and evaluate the success of an open-world ML approach, it is first critical to define \oodlong data. Existing work on open-world machine learning \cite{cvpr16_open_set_networks, dhamija2018agnostophobia, bendale2015openworld} defines an example $\bfx$ as \ood if it is drawn from a marginal distribution $P^{\text{out}}_{X}$ (over $\mathcal{X}$, the input feature space) which is different from $P^{\text{in}}_{X}$ and has a label set that is disjoint from that of in-distribution data. As a concrete example, consider a classifier trained on the CIFAR-10 image dataset \cite{krizhevsky2009learning}. This dataset only has 10 output classes and does not include classes for digits such as `3' or `7' like the MNIST \cite{lecun1998mnist} dataset or `mushroom' or `building' like the Imagenet dataset \cite{imagenet2009}. Thus, these datasets can act as a source of OOD data.

\subsubsection{OOD detectors} \label{subsubsec: back_ood_detect}
Here, we only review recent approaches to \ood detection that scale to DNNs used for image classification. Hendrycks and Gimpel \cite{hendrycks2017iclr} proposed a method for detecting OOD inputs for neural networks which uses a threshold for the output confidence vector to classify an input as in/out-distribution. This method relies on the assumption that the classifier will tend to have higher confidence values for in-distribution examples than OOD examples. An input is classified as being OOD if its output confidence value is smaller than a certain learned threshold.

In this work, we evaluate the state-of-the-art OOD detectors for DNNs proposed by Liang et al. \cite{liang2017ODIN} (ODIN) and Lee et al. \cite{lee2017training} which also use output thresholding for \ood detection but significantly improve upon the baseline approach of Hendrycks and Gimpel \cite{hendrycks2017iclr}. The ODIN detector uses temperature scaling and input pre-processing to improve detection rates. Lee et al. \cite{lee2017training} propose a modification to the training procedure to ensure that the neural network outputs a confidence vector which has probabilities uniformly distributed over classes for OOD inputs. However, as OOD inputs are unavailable at training time, they generate synthetic data,  using a modified Generative Adversarial Network (GAN), which lies on the boundary between classes to function as OOD data. \par

\noindent \textbf{Summary and implications:} Most previous work on adversarial examples has used data from the training or test datasets as starting points for the generation of adversarial examples. However, given the the importance of open-world machine learning for the deployment of machine learning systems in practice, this choice \emph{fails to consider an important attack vector}. On the other hand, previous work dealing with the open-world learning paradigm (Section \ref{subsec: back_open_world}) has largely not considered the possibility that this data could be adversarial, in addition to being OOD. In the next section, we demonstrate how we combine the two paradigms of open world learning and evasion attacks.


\section{Open-world Evasion Attacks}\label{sec:methods}
\begin{table*}[!ht]
    \renewcommand{\arraystretch}{1.4}
    \caption{Summary of results on CIFAR-10 trained models. Novel results and empirical conclusions from this paper are in \textbf{bold}. The last column shows the successful \oodAdvExamples for each defense, where each of them is generated to be classified as an $airplane$ by the network.
    }
    \label{tab: summary_table}
    \resizebox{\textwidth}{!}{%
        \begin{tabular}{lccccr}
            \toprule
            & & \multicolumn{3}{c}{Behavior on data type} & \\ \cline{4-5}
            Defense Type & Defense Name &  & \begin{tabular}[c]{@{}c@{}} In-distribution adversarial \\ (white-box, adaptive) \end{tabular}& \textbf{\begin{tabular}[c]{@{}c@{}} Out-of-distribution adversarial \\(white-box, adaptive) \end{tabular}} & \begin{tabular}[c]{@{}c@{}} Representative \\ \oodAdvExamples \end{tabular}\\ \midrule
            \multirow{4}{*}{\parbox[t]{2mm}{\rotatebox[origin=c]{0}{N.A.}}} & \multirow{4}{*}{Undefended} & & \multirow{4}{*}{\begin{tabular}[c]{@{}c@{}}Not robust \cite{madry_towards_2017,carlini2017towards} \\ Rate: 100.0, Conf: 1.00 \end{tabular}} & \multirow{4}{*}{\begin{tabular}[c]{@{}c@{}}Not robust\\ Rate: 100.0, Conf: 1.00 (ImageNet)\end{tabular}} & \multirow{4}{*}{\includegraphics[width=0.95in
            ]{images/org_table_2.png}}\\
            & & & \begin{tabular}[c]{@{}c@{}} \end{tabular} & \begin{tabular}[c]{@{}c@{}} \end{tabular} & \\
            & & & \begin{tabular}[c]{@{}c@{}} \end{tabular} & \begin{tabular}[c]{@{}c@{}} \end{tabular} & \\
            & & & \begin{tabular}[c]{@{}c@{}} \end{tabular} & \begin{tabular}[c]{@{}c@{}} \end{tabular} & \\ 
            \midrule
            \parbox[t]{2mm}{\multirow{2}{*}{\rotatebox[origin=c]{0}{\begin{tabular}[c]{@{}c@{}} Robust\\ Training \end{tabular}}}} & Iterative adv. training \cite{madry_towards_2017} & & \begin{tabular}[c]{@{}c@{}}Somewhat robust \cite{madry_towards_2017}\\ Rate: 22.9, Conf: 0.81 \end{tabular} & \begin{tabular}[c]{@{}c@{}}Not robust \\ {Rate: 87.9, Conf: 0.86 (Gaussian Noise)} \end{tabular} & \multirow{2}{*}{\includegraphics[width=0.95in
            ]{images/adv_train_table_2.png}}\\ 
            & Convex polytope relaxation \cite{kolter2017provable} & & \begin{tabular}[c]{@{}c@{}}Provably robust \cite{kolter2017provable}\\ Rate: 15.1, Conf: 0.41 \end{tabular} & \begin{tabular}[c]{@{}c@{}}Somewhat robust \\ Rate: 29.1, Conf: 0.32 (Gaussian Noise)\end{tabular} & \\ \midrule
            \parbox[t]{2mm}{\multirow{2}{*}{\rotatebox[origin=c]{0}{\begin{tabular}[c]{@{}c@{}} OOD\\ Detection \end{tabular}}}} & ODIN \cite{liang2017ODIN} & & N.A. & \begin{tabular}[c]{@{}c@{}}Not robust \\ Rate: 81.6, Conf: 0.97 (Internet Photographs)\end{tabular} & \multirow{2}{*}{\includegraphics[width=0.95in
            ]{images/detector_table_2.png}} \\
            & Confidence-calibrated \cite{lee2017training} & & N.A. & \begin{tabular}[c]{@{}c@{}} Somewhat robust \\ Rate: 47.1 , Conf: 0.99 (VOC12) \end{tabular} & \\ \bottomrule
        \end{tabular}
    }
\vspace{-10pt}
\end{table*}

Deployed ML systems must be able to robustly handle inputs which are drawn from distributions other than those used for training/testing. We thus define \emph{open-world evasion attacks}, which can use arbitrary points from the input space to generate \emph{out-of-distribution adversarial examples}, making our work the first to combine the paradigms of adversarial examples and open-world learning. We then analyze how they are effective in bypassing OOD detectors and defenses for in-distribution adversarial examples, including adversarial example detectors, making them a potent threat in the open-world machine learning model. These results are summarized in Table \ref{tab: summary_table}. Finally, we examine how robustness against OOD adversarial examples can be achieved.

\subsection{\oodAdvExamples}
In the open-world learning model, an adversary can generate adversarial examples using \ood data, and is not restricted to in-distribution data. In order to carry out an evasion attack in this setting, the adversary generates an \emph{OOD adversarial example} starting from $\bfx_{\text{\ood}}$.

\begin{definition}[\ood adversarial examples]
	An \ood adversarial example $\tilde{\bfx}_{\text{OOD}}$ is generated from an \ood example $\bfx_{\text{OOD}}$ drawn from $P_{X}^{\text{out}}$ by adding a perturbation $\bm{\delta}$ with the aim of inducing classification in a target class $T \in \mathcal{Y}_1$, i.e. $f(\tilde{\bfx}_{\text{OOD}}) = T$. 
\end{definition}

Any attack method used to generate adversarial examples starting from in-distribution data \cite{szegedy2014intriguing,goodfellow2014explaining,chen2017ead,athalye2018obfuscated}, even against defended classifiers, can be used to generate \ood adversarial examples. These attack methods are typically calibrated to ensure that adversarial examples are misclassified with high confidence (see Section \ref{sec:setup} on our design choices). Next, we highlight the importance of constructing OOD adversarial examples to fool classifiers by discussing the limitations of directly using unmodified/benign OOD data. 

\noindent \textbf{Limitations of unmodified \ood data for evasion attacks:}
While unmodified \ood data already represents a concern for the deployment of ML classifiers (Section \ref{subsec: back_open_world}), we now discuss why they are severely limited from an adversarial perspective. First, with unmodified OOD data, the typical output confidence values are small, while an attacker aims for high-confidence targeted misclassification. Second, the attacker will have no control over the target class reached by the unmodified \ood example. Finally, due to the low typical output confidence values of unmodified \ood examples they can easily be detected by state-of-the-art OOD detectors (Section \ref{subsubsec: back_ood_detect}) which rely on confidence thresholding. We provide quantitative evidence for this discussion by comparing the behavior of state-of-the-art classifiers on unmodified OOD data and OOD adversarial examples (details in Appendix \ref{appsec: benign_vs_adv}).

\subsubsection{Evading OOD detectors}
OOD detectors are an essential component of any open-world learning system and \ood adversarial examples should be able to bypass them in order to be successful. State-of-the-art OOD detectors mark inputs which have confidence values below a certain threshold as being OOD. The intuition is that a classifier is more likely to be highly confident on in-distribution data. Recall that when generating adversarial examples, the adversary aims to ensure high-confidence misclassification in the desired target class. The goal of an adversary seeking to generate high-confidence targeted OOD adversarial examples will align with that of an adversary aiming to bypass an OOD detector. In other words, \emph{OOD adversarial examples that achieve high-confidence targeted misclassification also bypass OOD detectors}. Our empirical results in Section \ref{subsec: ood_detection} demonstrate this conclusively, with OOD adversarial examples inducing high false negative rates in the OOD detectors, which mark them as being in-distribution.

\subsubsection{Evading robust training based defenses}
Robustly trained neural networks \cite{madry_towards_2017,kolter2017provable,raghunathan2018certified} (recall Section \ref{subsubsec: back_robust_training}), incorporate the attack strategy into the training process. Since the training and test data are drawn in an i.i.d. fashion, the resulting neural networks are robust at test time as well. However, these networks may not be able to provide robustness if the attack strategy were to be changed. In particular, we change the starting point for the generation of adversarial examples and since \emph{the training process for these robust defenses does not account for the possibility of OOD data being encountered at test time, they remain vulnerable to \oodAdvExamples}. We find that for defenses based on robust training, \oodAdvExamples are able to increase targeted success rate by $4\times$ (Sections \ref{subsubsec: ood_adv_trained} and \ref{subsubsec: ood_prov_train}). This finding illustrates the potent threat of open-world evasion attacks, which must be addressed for secure deployment of ML models in practice. We further demonstrate that adversarial example detectors such as MagNet and Feature Squeezing can be similarly bypassed by incorporating the metrics and pre-processing they use into the attack objective for \oodAdvExamples.

\subsubsection{Real-world attacks} 
Since the aim of the open-world threat model is to elucidate the wider range of possible threats to a deployed ML model than previously considered, we analyze the possibility of realizing \oodAdvExamples in the following real-world settings:
\begin{enumerate}
    \item \textbf{Physical attacks:} We consider attacks on a traffic sign recognition system where an adversary uses custom signs and logos in the environment as a source of OOD adversarial examples, since the classifier has only been trained on traffic signs. In a physical setting, there is the additional challenge of ensuring that the \oodAdvExamples remain adversarial in spite of environmental factors such as lighting and angle. We ensure this by incorporating random lighting, angle and re-sizing transformations into the OOD adversarial example generation process \cite{Athalye17,Sitawarin17,Evtimov17}.
    \item \textbf{Query-limited black-box attacks:} We use \ood adversarial examples to carry out a Denial of Service style attack on a content moderation model provided by Clarifai \cite{clarifai}, by \emph{classifying clearly unobjectionable content as objectionable with high confidence}. Since we only have query-access to the model being attacked, the model gradients usually needed to generate adversarial examples (see Section \ref{subsec: setup_attack}) have to be estimated. This is done using the finite difference method with random grouping based query-reduction \cite{bhagoji_eccv_2018}.
\end{enumerate}
Our results in Section \ref{subsec: real-world} show that \oodAdvExamples remain effective in these settings and are able to successfully attack content moderation and traffic sign recognition systems. 

\subsection{Towards Robust Open-World Deep Learning}
A robust open-world deep learning system is expected to satisfy the following two properties: (i) It should have high accuracy in detecting unmodified and adversarial \ood inputs; (ii) It should have high accuracy in classifying unmodified and adversarial in-distribution inputs. To move towards a robust open-world deep learning system, we take inspiration from previous work on selective prediction \cite{el2010backclass, chang1994backclass, geifman2017backclass, mccoyd2018background} which augments classifiers for in-distribution data with an additional class (referred to as background class) so they can be extended to open-world learning environment and detect OOD inputs. Further, since iterative adversarial training \cite{madry_towards_2017} enables the robust classification of in-distribution adversarial examples, we can intuit that a similar approach may provide robustness to \oodAdvExamples. Thus, we examine a \emph{hybrid approach where we use iterative adversarial training to ensure robust classification of \ood data, both unmodified and adversarial, to the background class}. Similar to other OOD detection approaches \cite{liang2017ODIN, lee2017training, hendrycks2017iclr, hendrycks2018outlierexpose}, selective prediction is semi-supervised, i.e. it assumes access to a small subset of \ood data at training time. We note that since all of
these state-of-the-art approaches consider the detection of specific (multiple) OOD datasets, we follow the same methodology for robust OOD classification. To achieve robust classification in the open-world environment, we perform iterative adversarial training with the following loss function:
\begin{equation}
    \tilde{\ell}_{g}(\mathbf{x_\text{in}},\mathbf{x_\text{OOD}},y) = \alpha \tilde{\ell}_{g}(\mathbf{x_\text{in}},y)+ (1 - \alpha)\tilde{\ell}_{g}(\mathbf{x_\text{OOD}}, y_b)
\end{equation}
where $\mathbf{x_\text{in}}\in P^{\text{in}}_{X},\mathbf{x_\text{OOD}}\in P^{\text{out}}_{X}$, y is true label for sample $\mathbf{x_{in}}$ and $y_b$ is the background class. $\tilde \ell_{g}(\cdot,\cdot)$ refers to the robust loss used in adversarial training (Eq. \ref{eq: adv_train}).

The question now arises: to what extent does this formulation satisfy the two desired properties from a robust open-world learning system? In particular, we examine if the following goals are feasible using small subsets of \ood data: i) robust classification of a single \ood dataset?, ii) generalization of robustness to multiple OOD datasets while training with a single one?, iii) simultaneous robustness to multiple OOD datasets while training with data from all of them? Again, we emphasize that these must be achieved while maintaining high accuracy on in-distribution data.

Our evaluation in Section~\ref{sec:defense} answers these questions in the affirmative. For example, we observe that a subset as small as \textit{0.5\%} of the total number of samples from an \ood dataset can significantly enhance robustness against OOD adversarial examples.

\section{Design Choices for Open-World Evasion Attacks}\label{sec:setup}
In this section, we present and examine the design choices we make to carry out our experiments on both evaluation and training of classifiers in the open-world model. In particular, we discuss the types of datasets, attack methods, models and metrics we consider. 

\subsection{Datasets}\label{sec:setup_datasets}
We consider 3 publicly available datasets for image classification as sources of in-distribution data for training ($P^{\text{in}}_{X,Y}$). These are MNIST \cite{lecun1998mnist}, CIFAR-10 \cite{krizhevsky2014cifar}, and ImageNet (ILSVRC12 release) \cite{imagenet2009}. When one of the above datasets is used for training, all the other datasets we consider are used as a source of \ood data. We consider two types of \ood data: i) semantically meaningful OOD data and ii) noise OOD data.

\noindent \textbf{Semantically meaningful OOD data:}  
Datasets such as MNIST, CIFAR-10 and Imagenet are \emph{semantically meaningful} as the images they contain generally have concepts recognizable to humans. To further explore the space of semantically meaningful \ood data, we also consider the VOC12 \cite{pascal-voc-2012} dataset as a source of \ood data. Further, to emulate the type of data that may be encountered in an open-world setting, we construct a \intphoto dataset by gathering 10,000 natural images from the internet using the Picsum service \cite{picsum}. To avoid any ambiguity over examples from different datasets that contain similar semantic information, we ensure the label set for semantically meaningful \ood examples is distinct from that of the in-distribution dataset. Additional details of these datasets are in Appendix~\ref{sec:appendix_dataset_details}.

\noindent \textbf{Noise \ood data:} By definition, \ood data does not have to contain recognizable concepts. Thus, we construct a Gaussian Noise dataset consisting of 10,000 images for each of which the pixel values are sampled from a Gaussian distribution. This dataset is very different from the image datasets considered before as it consists of random points in the input space which have no relation to any naturally occuring images. In settings where inputs to an ML classifier are not checked for any sort of semantics, this dataset is a viable input and thus must be analyzed when checking for robustness.  

\subsection{OOD Evasion Attack Methods}\label{subsec: setup_attack}
Iterative optimization based methods have been found to be the most effective for solving the relaxed version of Eq. \ref{eq: comb} \cite{szegedy2014intriguing,goodfellow2014explaining,carlini2017towards,madry_towards_2017,athalye2018obfuscated,moosavi2015deepfool} to generate successful in-distribution adversarial examples regardless of the choice of distance function \cite{madry_towards_2017,athalye2018obfuscated}. Thus, we exclusively use these to generate \ood adversarial examples. 

\subsubsection{Loss functions and optimizers} 
We use two commonly used loss functions $\ell_g(\cdot,T)$ to generate \ood adversarial examples. The first is the standard \emph{cross-entropy loss} \cite{goodfellow2016deep}. The second is the \emph{logit loss} from Carlini and Wagner \cite{carlini2017towards}:
\begin{align}
\ell^{\text{adv}}_g(\xad_\text{OOD},T)=\max(\max\{\bfZ_g(\xad_\text{OOD})_i: i \neq T\}-\bfZ_g(\xad_\text{OOD})_T, -\kappa),
\end{align}
where $\kappa$ is a confidence parameter that can be adjusted to control the strength of the adversarial example.

\noindent \textbf{Optimizer choice:} The loss $\ell^{\text{adv}}_{g}(\bfx_\text{OOD},T)$ is minimized using \emph{Projected Gradient Descent} (PGD) which we choose for its state-of-the-art attack performance \cite{madry_towards_2017,athalye2018obfuscated}. PGD iteratively minimizes the loss while projecting the obtained minimizer onto the constraint set $\mathcal{H}$ (subset of $\mathcal{X}$ depending on $d(\cdot,\cdot)$) for $t$ iterations,
\begin{align}
\label{eq: pgd}
\xad^{t}_\text{OOD} = \Pi_{\mathcal{H}}(\xad^{t-1}_\text{OOD} - \alpha \cdot \text{sign}(\nabla_{\xad^{t-1}}\ell^{\text{adv}}_{g}(\xad^{t-1}_\text{OOD}, T))),
\end{align}
where $\Pi$ is a projection operator, $\xad^0_\text{OOD}=\bfx_\text{OOD}$ and $\xad^{t}_\text{OOD}$ is the final adversarial example. This formulation allows us to work with fixed constraint sets $\mathcal{H}$, which we use throughout to compare across datasets, attacks and defenses. As noted by Athalye et al. \cite{athalye2018obfuscated}, the exact choice of the optimizer does not matter much in practice, as long as an iterative optimization procedure is used. We refer to the use of PGD with the cross-entropy loss as \pgdx \xspace and with the logit loss from Carlini and Wagner as \pgdcw. 

We run PGD for $100$-$1000$ iterations for every attack. Attacks to generate \ood adversarial examples usually converge within 100 iterations. If they do not converge in 100 iterations, we monitor the loss and stop if it has plateaued. The step size is a hyper-parameter that is varied in each experiment depending on the maximum possible perturbation and attack performance.

\noindent \textit{Note.}  While most of our experiments are for adaptive white-box adversaries, in Section \ref{subsec: real-world} we use query-based black-box attacks which rely on PGD but due to a lack of access to the true gradient, we use estimated gradients instead.

\subsubsection{Choosing targets for attacks} 
We only consider \emph{targeted} attacks for two reasons. First, they are strictly harder than non-targeted attacks for the adversary \cite{carlini2017towards}. Second, unmodified OOD examples have no ground truth labels, which raises difficulties in defining untargeted attacks and comparing them to the in-distribution case. We use two methods for target label selection for our attacks: i) \emph{random}: this refers to the selection of a random target label from a set that excludes the current predicted label for an unmodified example (referred to as `rand'); ii) \emph{least likely}: this refers to the selection of the label with the least value in the output confidence vector for a given input (referred to as `LL'). 


\subsubsection{Overcoming gradient obfuscation} 
Defenses such as feature squeezing \cite{xu2017feature} obfuscate gradients, which can cause PGD-based attacks to fail due to the presence of uninformative gradients \cite{athalye2018obfuscated, uesato2018adversarial}. In particular, Athalye et al. \cite{athalye2018obfuscated} demonstrated that these obfuscated gradients are caused simply due to complex input pre-processing which can be approximated by simpler, differentiable transformations through which gradients are able to pass. We adopt this approach for our experiments in Section \ref{subsec: adv_detectors}.

\subsubsection{Distance constraints}
In this paper, we use the $L_{\infty}$ perturbation constraint for most of our attacks, except for the attack on feature squeezing \cite{xu2017feature}, where the $L_{\infty}$ metric cannot be used due to bit depth reduction and thus the $L_{2}$ perturbation is adopted instead. These metrics are widely used to generate in-distribution adversarial examples because examples $\bfx$ and $\xad$ that are $\epsilon$-close in the these distance metrics, are also visually similar \cite{szegedy2014intriguing,goodfellow2014explaining,carlini2017towards}. For any vector $\bfz \in \mathbb{R}^n$, these metrics are:- i) $L_{\infty}:\|z\|_{\infty}=\max_{i \in [n]} |\bfz_i|$, i.e. the absolute value of the largest component of the vector; ii) $L_2:\|z\|_2=\sqrt{\sum_{i=1}^{n}\bfz_i^2}$, i.e. the square root of the sum of the squared value of the components.

\noindent \textbf{Why use distance constraints for \ood adversarial examples?:} There are two main reasons why we use distance constraints to generate \ood adversarial examples. The first reason, which \emph{applies only to semantically meaningful \ood data}, is to model the content in the input that the adversary wishes to preserve, in spite of it being \ood. In other words, the starting point itself models a constraint on the adversary, which may arise from the environment (see Section \ref{subsubsec: traffic_sign_attack} for an example for traffic sign recognition systems) or to prevent the \ood adversarial example from having undesirable artifacts, e.g. turning a non-objectionable image into an objectionable one (see Section \ref{subsubsec: content_mod_attack}).

The second reason, which applies to both semantically meaningful and noise \ood data, stems from the \emph{need to measure the spread of successful \ood adversarial examples in the input space}. Previous work has measured the spread of adversarial examples around the training and test data, in terms of $L_p$ distance constraints and found that for undefended models adversarial examples are present close to their starting points. On the other hand, while the use of robust training defenses makes it challenging to find adversarial examples within a given constraint set, we show for \ood data, successful \ood adversarial examples can be found in small $L_p$ balls around unmodified starting points for both undefended and defended models. We note that for noise \ood data it is possible to relax distance constraints to generate \ood adversarial examples which will lead to higher attack success rates. In Section \ref{subsubsec: traffic_sign_attack}, we demonstrate open-world evasion attacks using custom signs on traffic sign recognition systems that do not restrict the perturbation budget as well.


\subsection{Metrics and Models}\label{sec:setup_metrics}

\subsubsection{Evaluation metrics} 
We consider the following metrics to measure the performance and robustness of image classifiers:\\
\noindent \textbf{Target success rate}: This is the percentage of adversarial examples classified as the desired target, which measures model robustness.\\
\noindent \textbf{Classification accuracy}: This is the percentage of in-distribution test data for which the predicted label matches the ground-truth label. It is not reported for OOD data as it has no ground truth labels.\\
\noindent \textbf{Mean classification confidence}: This is the mean value of the output probability corresponding to the predicted label of the classifier on correctly classified inputs in the case of benign in-distribution data. For adversarial examples, both in-distribution and OOD, it is the mean value of the output probability corresponding to the target label for successful adversarial examples. The confidence values lie in $[0,1]$.

\subsubsection{Models} 
We experiment with three robust training defenses (Iterative adversarial training~\cite{madry_towards_2017}, Adversarial logit pairing~\cite{kannan2018adversarial}, and robust training with convex outer polytope~\cite{kolter2017provable}), two adversarial example detectors (Feature Squeezing~\cite{xu2017feature} and MagNet~\cite{meng2017magnet}), and two OOD detectors (ODIN~\cite{liang2017ODIN} and Confidence calibrated classifiers~\cite{lee2017training}). These works use multiple different neural network models, each trained on the MNIST, CIFAR-10, or ImageNet datasets. We show the architecture and performance of the 11 DNNs from these works used in this paper in Table \ref{table:dataset_and_model_details}. We use multiple models for each dataset to carefully compare with previous work. All the DNNs we consider are convolutional neural networks as these lead to the state-of-the-art performance on image classification \cite{image_benchmarks}. To characterize performance on in-distribution data, we report both the classification accuracy and mean classification confidence on these models. Following the convention in previous work \cite{madry_towards_2017, Carlini16, kolter2017provable}, we report the perturbation budget for models trained on MNIST dataset using [0,1] scale instead of [0,255].

All experiments are run on a GPU cluster with 8 Nvidia P100 GPUs using mainly \textsf{Python}-based \textsf{Tensorflow} \cite{tensorflow2015-whitepaper} to implement and run DNNs along with the \textsf{Numpy} and \textsf{Scipy} packages for other mathematical operations. We also use \textsf{Pytorch} \cite{paszke2017automatic} in cases when previous work that we build on has used it. All our code and data will be publicly released for the purposes of reproducible research.

\begin{table}[t]
\fontsize{15pt}{15pt}\selectfont
\renewcommand{\arraystretch}{1.2} 
\caption{Deep neural networks used for each dataset in this work. The top-1 accuracy and confidence values are calculated using the test set of the respective datasets.}
\label{table:dataset_and_model_details}
\resizebox{0.48\textwidth}{!}{%
\begin{threeparttable}
\begin{tabular}{cccc}
\toprule
Dataset & Model & \begin{tabular}[c]{@{}c@{}}Classification \\ accuracy ($\%$)\end{tabular} & \begin{tabular}[c]{@{}c@{}}Mean \\ confidence \end{tabular} \\
\midrule
\multirow{3}{*}{MNIST~\cite{lecun1998mnist}} & 4-layer CNN($M_1$)~\cite{madry_towards_2017}\tnote{*} & 98.8 & 0.98 \\
& 4-layer CNN ($M_2$)~\cite{kolter2017provable}\tnote{\textdagger} & 98.2 & 0.98 \\
& 7-layer CNN ($M_3$)~\cite{carlini2017towards} & 99.4 & 0.99\\  \midrule
\multirow{4}{*}{CIFAR-10~\cite{krizhevsky2014cifar}} & \begin{tabular}[c]{@{}c@{}}Wide Residual net \\ (WRN-28-10)~\cite{zagoruyko2016wide}\end{tabular} & 95.1 & 0.98 \\
& WRN-28-10-A~\cite{zagoruyko2016wide, madry_towards_2017}\tnote{*} & 87.2 & 0.93 \\
& WRN-28-1~\cite{wong2018scaling}\tnote{\textdagger} & 66.2 & 0.57 \\ 
& DenseNet~\cite{HuangLW16densenet}& 95.2 & 0.98\\
& VGG13~\cite{simonyan2014vgg} & 80.1 & 0.94 \\ 
& All Convolution Net \cite{springenberg2014striving} & 85.7 & 0.96 \\
\midrule
\multirow{2}{*}{ImageNet~\cite{imagenet2009}} & MobileNet~\cite{Howard17mobilenet} & 70.4 & 0.71 \\
 & \begin{tabular}[c]{@{}c@{}}ResNet-v2-50~\cite{kannan2018adversarial,he2016resnet}\tnote{\textdaggerdbl}\\ (for 64$\times$64 size images)\end{tabular} & 60.5 & 0.28 \\ \bottomrule
\end{tabular}
\begin{tablenotes}
\item[*] Iterative adversarial training from Madry et. al.~\cite{madry_towards_2017}.
\item[\textdagger] Robust training with convex polytope relaxation from Wong et. al.~\cite{wong2018scaling}.
\item[\textdaggerdbl] Adversarial logit pairing (ALP) from Kannan et. al.~\cite{kannan2018adversarial}.
\end{tablenotes}
\end{threeparttable}
}
\vspace{-5pt}
\end{table}

\section{Results}\label{sec:results}
In this section, we present the experimental results. We first evaluate the robustness of OOD detectors to \oodAdvExamples. Next, we analyze the performance of state-of-the-art defenses against adversarial examples in the open-world learning model. We also evaluate adversarial detectors and secondary defenses in this setup. Finally, we demonstrate the relevance of open-world evasion attacks with real-world attacks using \oodAdvExamples.

\subsection{OOD detectors are not robust}\label{subsec: ood_detection}

In open-world deep learning, the first step is to detect \oodlong inputs. Previous work has demonstrated that unmodified OOD inputs to a DNN can be detected with high probability by using an additional detector \cite{liang2017ODIN,lee2017training} (recall Section \ref{subsubsec: back_ood_detect}). In this section, we evaluate the robustness of these OOD detection approaches to \oodAdvExamples. Specifically, we evaluate two state-of-the-art OOD detection methods, ODIN \cite{liang2017ODIN} and confidence calibrated classification \cite{lee2017training}, on trained CIFAR-10 models. \par
%
\noindent \textbf{OOD detector setup.} The success of an OOD detector is measured by False Negative Rate (FNR), which represents the fraction of OOD inputs the detector fails to detect. The threshold values reported in \cite{liang2017ODIN, lee2017training} are calibrated such that the True Negative Rate (TNR) i.e., the fraction of in-distribution inputs the detector classifies as non-OOD, is equal to 95$\%$. For both detectors, the \pgdx~attack with 100 iterations and random target label selection is used.

\noindent \textbf{Summary of results.} While OOD detectors achieve good detection performance on unmodified OOD data, \emph{these  detectors can be evaded with \oodAdvExamples}. The key reason for the failure to detect \oodAdvExamples is that both ODIN \cite{liang2017ODIN} and confidence calibrated classification \cite{lee2017training} rely on low output confidence values to distinguish OOD examples. As we will discuss next in Section~\ref{subsec: strong_defenses} (Table~\ref{table:default_robust_defenses}), our results show that \oodAdvExamples can achieve target output classification with confidence equal to (and often higher than) that of in-distribution images.


\subsubsection{Effect of OOD attacks on ODIN} We use the code and pre-trained DenseNet \cite{HuangLW16densenet} model on CIFAR-10 dataset from Liang et al. \cite{liang2017ODIN}. For consistency, we follow Liang et al. \cite{liang2017ODIN} and use the temperature scaling and perturbation budget for input pre-processing equal to 1000 and 0.0014 respectively.
%

\noindent \textbf{\oodAdvExamples can evade the ODIN detector successfully.} We first test the performance of ODIN with multiple unmodified OOD datasets. As expected, ODIN achieves more than 78$\%$ detection accuracy for all unmodified OOD datasets. However, \emph{the detection rate of ODIN drastically decreases with \oodAdvExamples} (Table \ref{table:ood_detectors}). For \oodAdvExamples generated with ImageNet dataset and $\epsilon$ of 16, ODIN misses 97.4$\%$ inputs. Except for Gaussian noise, the mean success rate is 98.3$\%$ for other OOD datasets with $\epsilon=16$. For \oodAdvExamples generated from Gaussian noise, we observe that the success in evading ODIN is dependent on the neural network model. As highlighted in Table~\ref{table:ood_detectors}, a DenseNet model leads to poor success of \oodAdvExamples from Gaussian noise. However, for a WRN-28-10 model, we found that the \oodAdvExamples generated from Gaussian noise achieve FNR close to 100$\%$.\par 

\begin{table}[!tbp] 
\caption{False Negative rate (FNR) of ODIN~\cite{liang2017ODIN} and confidence calibrated classifier~\cite{lee2017training} approaches for \oodAdvExamples. The results are reported with the respective models for each detector trained on CIFAR-10 dataset. The TNR of each detector with in-distribution dataset is 95$\%$. These results show that a high percentage of \oodAdvExamples can evade OOD detectors. The maximum values among all OOD datasets are highlighted in bold.}
\label{table:ood_detectors}
\resizebox{0.48\textwidth}{!}{%
\begin{tabular}{ccccccc}
\toprule
OOD dataset &  & \multicolumn{2}{c}{ODIN~\cite{liang2017ODIN}} &  & \multicolumn{2}{c}{\begin{tabular}[c]{@{}c@{}}Confidence-calibrated\\ classifier~\cite{lee2017training}\end{tabular}} \\ \cline{3-4} \cline{6-7}
 &  & $\epsilon$ = 8.0 & $\epsilon$ = 16.0 &  & $\epsilon$ = 8.0 & $\epsilon$ = 16.0 \\ \midrule
ImageNet &  & 68.8 & 97.4 &  & 46.4 & \textbf{47.5} \\
VOC12 &  & 74.4 & 97.4 &  & \textbf{47.1} & 47.2 \\
\begin{tabular}[c]{@{}c@{}}Internet\\ Photographs\end{tabular} &  & \textbf{81.6} & 98.7 &  & 42.5 & 45.4 \\
MNIST &  & 72.6 & \textbf{99.8} &  & 4.6 & 5.2 \\
\begin{tabular}[c]{@{}c@{}}Gaussian \\ Noise\end{tabular} &  & 0 & 4.2 &  & 20.9 & 21.9 \\ \bottomrule
\end{tabular}
}
\vspace{-15pt}
\end{table}
\subsubsection{Effect of OOD attacks on confidence calibrated classifiers} For consistency with prior work of Lee et al.~\cite{lee2017training}, we used a similar model (VGG13) and training procedure. We also validate the results from Lee et al.~\cite{lee2017training} by evaluating this detector on unmodified OOD datasets. \par 
%
\noindent \textbf{Up to 47.5$\%$ \oodAdvExamples could bypass the detection approach based on confidence calibrated classifiers.} In our baseline experiments with unmodified OOD datasets, we found that the confidence-calibrated classifier has good detection performance. For example, unmodified ImageNet and VOC12 dataset are correctly detected with an accuracy of $83.5\%$ and $88.3\%$, respectively. However, the detection performance degrades significantly for adversarial examples generated from OOD datasets except MNIST (Table~\ref{table:ood_detectors}). For example, when $\epsilon$ equals to 16, more than $45\%$ adversarial examples generated from ImageNet, VOC12, and Internet Photographs datasets are missed by the detector. \par 

However, in comparison to ODIN \cite{liang2017ODIN}, the gradient-based attacks for this detector fail to achieve close to 100$\%$ FNR. We observe that even with an \emph{unconstrained} adversarial attack the FNR doesn't approach 100$\%$. We speculate that this behavior might be due to non-informative gradients presented by the model at the input. It should be noted that first-order attack approaches which can succeed in presence of obfuscated gradients \cite{athalye2018obfuscated} aren't applicable here. This is because instead of any additional input-processing step the gradients are obfuscated by the model itself. 

\subsection{Fooling robustly trained models}\label{subsec: strong_defenses}
In this section, we first evaluate the robustness of baseline, undefended models (trained with natural training) to \oodAdvExamples. Next, we evaluate robustly trained models with \oodAdvExamples. For robust training, we consider the two state-of-the-art approaches discussed in Section \ref{subsubsec: back_robust_training} which are iterative adversarial training \cite{madry_towards_2017} and robust training using the convex polytope relaxation \cite{kolter2017provable}. For \emph{undefended models}, we show that \oodAdvExamples achieve target success rates almost identical to adversarial examples generated from in-distribution data. We further show that \emph{robustly trained models} are much less robust against \oodAdvExamples. 

\begin{figure}[!htb]
	\centering
	\begin{subfigure}[t]{\linewidth}
		\centering
		\includegraphics[width=\linewidth]{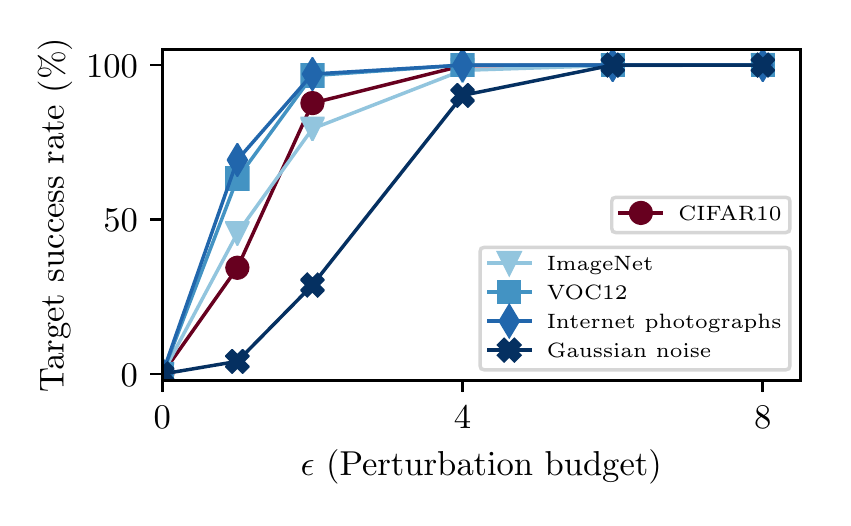}
		\caption{Lack of robustness of natural training.}
		\label{fig:cifar_clean_pgd}
	\end{subfigure}
	\begin{subfigure}[t]{\linewidth}
		\centering
		\includegraphics[width=\linewidth]{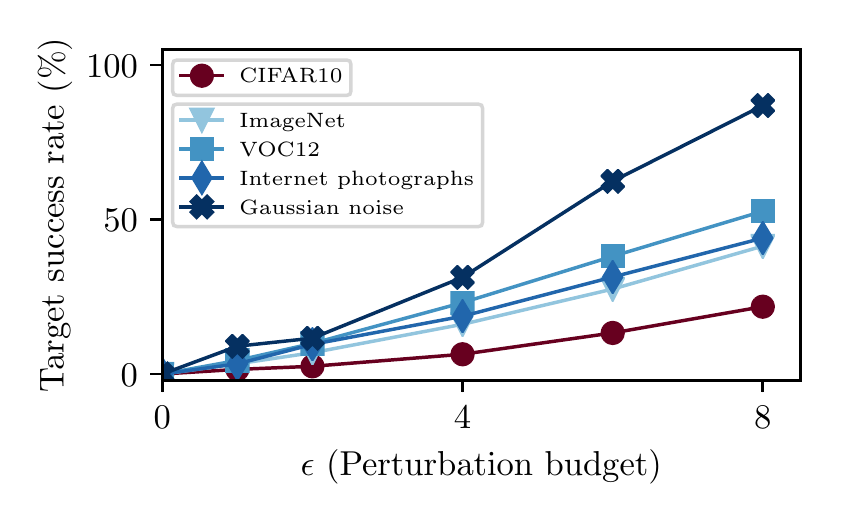}
		\caption{{Lack of robustness of iterative adversarial training \cite{madry_towards_2017}.}}
		\label{fig:cifar_adv_trained_pgd}
	\end{subfigure}
	\caption{Target success rate of adversarial examples generated from different datasets for the state-of-the-art WRN-28-10 \cite{zagoruyko2016wide} model trained on CIFAR-10 \cite{krizhevsky2014cifar}. The \textsf{PGD-xent} attack is used (Section \ref{subsec: setup_attack}) with $\epsilon$ ($l_{\infty}$ perturbation budget) up to 8 and random target label selection. Though iterative adversarial training \cite{madry_towards_2017} (with $\epsilon=8$) improves robustness for in-distribution data (CIFAR-10), \oodAdvExamples are up to $4 \times$ as successful as those generated from in-distribution data.} 
	\label{fig:cifar_adv_trained_comparison}
	\vspace{-10pt}
\end{figure}




\begin{table*}[!ht]
\fontsize{15pt}{15pt}\selectfont
\renewcommand{\arraystretch}{1.4} 
\caption{Target success rate of adversarial examples generated from different datasets for models trained with iterative adversarial training~\cite{madry_towards_2017}, adversarial logit pairing (ALP) \cite{kannan2018adversarial}, and convex polytope relaxation~\cite{kolter2017provable}. The $l_\infty$ norm used to generate adversarial examples is listed along with the training dataset. The maximum target success by the adversarial examples for every model is highlighted in bold. The results for in-distribution data are highlighted in \textit{italics}.  
}
\label{table:default_robust_defenses}
\resizebox{\textwidth}{!}{%
\begin{tabular}{ccccccccccccccccccccc}
\toprule
\multicolumn{3}{c}{} && \multicolumn{6}{c}{\begin{tabular}[c]{@{}c@{}}Iterative adversarial\\ training~\cite{madry_towards_2017}\end{tabular}} && \multicolumn{3}{c}{\begin{tabular}[c]{@{}c@{}}Adversarial\\ logits pairing~\cite{kannan2018adversarial}\end{tabular}} && \multicolumn{6}{c}{\begin{tabular}[c]{@{}c@{}}Provable \\ Defenses~\cite{kolter2017provable}\end{tabular}} \\ \cmidrule{5-10} \cmidrule{12-14} \cmidrule{16-21} 
\multicolumn{1}{c}{\begin{tabular}[c]{@{}c@{}}Test ($\downarrow$)~\textbackslash{}~Train ($\rightarrow$)\\ dataset\end{tabular}} & & &&\multicolumn{3}{c}{MNIST ($\epsilon = 0.3$)} & \multicolumn{3}{c}{CIFAR-10 ($\epsilon = 8$)} && \multicolumn{3}{c}{ImageNet ($\epsilon = 16$)} && \multicolumn{3}{c}{MNIST ($\epsilon = 0.1$)} & \multicolumn{3}{c}{CIFAR-10 ($\epsilon = 8$)} \\ \cmidrule{5-10} \cmidrule{12-14} \cmidrule{16-21}

\multicolumn{1}{c}{} & \multirow{2}{*}{Attack} & \multirow{2}{*}{\begin{tabular}[c]{@{}c@{}}Target \\ labels\end{tabular}} && \multicolumn{1}{c}{\multirow{2}{*}{\begin{tabular}[c]{@{}c@{}}success \\ rate ($\%$)\end{tabular}}} & \multicolumn{2}{c}{confidence} & 
\multicolumn{1}{c}{\multirow{2}{*}{\begin{tabular}[c]{@{}c@{}}success \\ rate ($\%$)\end{tabular}}}& \multicolumn{2}{c}{confidence} && 
\multicolumn{1}{c}{\multirow{2}{*}{\begin{tabular}[c]{@{}c@{}}success \\ rate ($\%$)\end{tabular}}} & \multicolumn{2}{c}{confidence} && 
\multicolumn{1}{c}{\multirow{2}{*}{\begin{tabular}[c]{@{}c@{}}success \\ rate ($\%$)\end{tabular}}} & \multicolumn{2}{c}{confidence} & 
\multicolumn{1}{c}{\multirow{2}{*}{\begin{tabular}[c]{@{}c@{}}success \\ rate($\%$) \end{tabular}}}& \multicolumn{2}{c}{confidence} \\ 
\cmidrule{6-7}  \cmidrule{9-10} \cmidrule{13-14} \cmidrule{17-18} \cmidrule{20-21} 

\multicolumn{3}{c}{} && \multicolumn{1}{c}{} & \multicolumn{1}{c}{clean} & \multicolumn{1}{c}{adv} & \multicolumn{1}{c}{} & \multicolumn{1}{c}{clean} &
\multicolumn{1}{c}{adv} && \multicolumn{1}{c}{} & \multicolumn{1}{c}{clean} &
\multicolumn{1}{c}{adv} && \multicolumn{1}{c}{} & \multicolumn{1}{c}{clean} &
\multicolumn{1}{c}{adv} & \multicolumn{1}{c}{} & \multicolumn{1}{c}{clean} 
& \multicolumn{1}{c}{adv}\\ \midrule

\multicolumn{1}{c}{\multirow{4}{*}{MNIST}} & \multicolumn{1}{c}{\multirow{2}{*}{\pgdx}} & \multicolumn{1}{c}{rand}  && \multicolumn{1}{c}{\textit{1.5}} & \multicolumn{1}{c}{\multirow{4}{*}{\textit{0.98}}} & \multicolumn{1}{c}{\textit{0.76}} & \multicolumn{1}{c}{5.1} & \multicolumn{1}{c}{\multirow{4}{*}{0.81}} & \multicolumn{1}{c}{0.60} && \multicolumn{1}{c}{98.8} & \multicolumn{1}{c}{\multirow{4}{*}{0.27}} & \multicolumn{1}{c}{0.96} && \multicolumn{1}{c}{\textit{0.6}} & \multicolumn{1}{c}{\multirow{4}{*}{\textit{0.97}}} & \multicolumn{1}{c}{\textit{0.64}}  & \multicolumn{1}{c}{3.9} & \multicolumn{1}{c}{\multirow{4}{*}{0.48}} & \multicolumn{1}{c}{0.37}\\ 

\multicolumn{1}{c}{} & \multicolumn{1}{c}{} & \multicolumn{1}{c}{LL}  && \multicolumn{1}{c}{\textit{0.0}} & \multicolumn{1}{c}{} & \multicolumn{1}{c}{\textit{0.0}} & \multicolumn{1}{c}{0.0} & \multicolumn{1}{c}{} &  \multicolumn{1}{c}{0.0} && \multicolumn{1}{c}{99.4} &  \multicolumn{1}{c}{} & \multicolumn{1}{c}{0.95} && \multicolumn{1}{c}{\textit{0.0}} & \multicolumn{1}{c}{} &  \multicolumn{1}{c}{\textit{0.0}} & \multicolumn{1}{c}{0} & \multicolumn{1}{c}{} & \multicolumn{1}{c}{0}\\ 

\multicolumn{1}{c}{} & \multicolumn{1}{c}{\multirow{2}{*}{\pgdcw}} & \multicolumn{1}{c}{rand} && \multicolumn{1}{c}{\textit{1.9}} & \multicolumn{1}{c}{} &  \multicolumn{1}{c}{\textit{0.71}} & \multicolumn{1}{c}{5.7} & \multicolumn{1}{c}{} &  \multicolumn{1}{c}{0.54}  && \multicolumn{1}{c}{97.0} & \multicolumn{1}{c}{} &   \multicolumn{1}{c}{0.91} && \multicolumn{1}{c}{\textit{1.2}} & \multicolumn{1}{c}{} &  \multicolumn{1}{c}{\textit{0.67}} & \multicolumn{1}{c}{5.1} & \multicolumn{1}{c}{} & \multicolumn{1}{c}{0.33}\\ 

\multicolumn{1}{c}{} & \multicolumn{1}{c}{} & \multicolumn{1}{c}{LL} && \multicolumn{1}{c}{\textit{0.0}} & \multicolumn{1}{c}{} &  \multicolumn{1}{c}{\textit{0.0}} & \multicolumn{1}{c}{0.0} & \multicolumn{1}{c}{} &  \multicolumn{1}{c}{0.0} && \multicolumn{1}{c}{96.6} & \multicolumn{1}{c}{} &  \multicolumn{1}{c}{0.88} && \multicolumn{1}{c}{\textit{0.0}} & \multicolumn{1}{c}{} &  \multicolumn{1}{c}{\textit{0.0}} & \multicolumn{1}{c}{0} & \multicolumn{1}{c}{} & \multicolumn{1}{c}{0}\\ \midrule

\multicolumn{1}{c}{\multirow{4}{*}{CIFAR-10}} & \multicolumn{1}{c}{\multirow{2}{*}{\pgdx}} & \multicolumn{1}{c}{rand} && \multicolumn{1}{c}{97.6} & \multicolumn{1}{c}{\multirow{4}{*}{0.77}} & \multicolumn{1}{c}{0.99} & \multicolumn{1}{c}{\textit{22.9}} & \multicolumn{1}{c}{\multirow{4}{*}{\textit{0.93}}} & \multicolumn{1}{c}{\textit{0.81}} && \multicolumn{1}{c}{\textbf{100.0}} & \multicolumn{1}{c}{\multirow{4}{*}{0.14}} & \multicolumn{1}{c}{0.99} && \multicolumn{1}{c}{67.2} & \multicolumn{1}{c}{\multirow{4}{*}{0.88}} & \multicolumn{1}{c}{0.97} & \multicolumn{1}{c}{\textit{15.1}} & \multicolumn{1}{c}{\multirow{4}{*}{\textit{0.27}}} & \multicolumn{1}{c}{\textit{0.41}}  \\ 

\multicolumn{1}{c}{} & \multicolumn{1}{c}{} & \multicolumn{1}{c}{LL}  && \multicolumn{1}{c}{95.3} & \multicolumn{1}{c}{} &  \multicolumn{1}{c}{0.99} & \multicolumn{1}{c}{\textit{5.1}} & \multicolumn{1}{c}{} &  \multicolumn{1}{c}{\textit{0.69}}  && \multicolumn{1}{c}{100.0} &  \multicolumn{1}{c}{} & \multicolumn{1}{c}{0.99} && \multicolumn{1}{c}{30.3} & \multicolumn{1}{c}{} &  \multicolumn{1}{c}{0.93} & \multicolumn{1}{c}{\textit{0.4}} & \multicolumn{1}{c}{} & \multicolumn{1}{c}{\textit{0.15}}\\ 

\multicolumn{1}{c}{} & \multicolumn{1}{c}{\multirow{2}{*}{\pgdcw}} & \multicolumn{1}{c}{rand} && \multicolumn{1}{c}{97.6} & \multicolumn{1}{c}{} &  \multicolumn{1}{c}{0.99} & \multicolumn{1}{c}{\textit{22.3}} & \multicolumn{1}{c}{} &  \multicolumn{1}{c}{\textit{0.76}} && \multicolumn{1}{c}{99.9} & \multicolumn{1}{c}{} &  \multicolumn{1}{c}{0.96} && \multicolumn{1}{c}{61.5} & \multicolumn{1}{c}{} & \multicolumn{1}{c}{0.97} & \multicolumn{1}{c}{\textit{16.4}} & \multicolumn{1}{c}{} & \multicolumn{1}{c}{\textit{0.35}}\\ 

\multicolumn{1}{c}{} & \multicolumn{1}{c}{} & \multicolumn{1}{c}{LL} && \multicolumn{1}{c}{93.8} & \multicolumn{1}{c}{} &  \multicolumn{1}{c}{0.98} & \multicolumn{1}{c}{\textit{4.4}} & \multicolumn{1}{c}{} &  \multicolumn{1}{c}{\textit{0.60}} && \multicolumn{1}{c}{99.5} & \multicolumn{1}{c}{} &  \multicolumn{1}{c}{0.95} && \multicolumn{1}{c}{27.6} & \multicolumn{1}{c}{} &  \multicolumn{1}{c}{0.91} & \multicolumn{1}{c}{\textit{0.4}} & \multicolumn{1}{c}{} & \multicolumn{1}{c}{\textit{0.15}}\\ \midrule

\multicolumn{1}{c}{\multirow{4}{*}{ImageNet}} & \multicolumn{1}{c}{\multirow{2}{*}{\pgdx}} & \multicolumn{1}{c}{rand} & & \multicolumn{1}{c}{97.2} & \multicolumn{1}{c}{\multirow{4}{*}{0.79}} &  \multicolumn{1}{c}{0.99} & \multicolumn{1}{c}{44.9} & \multicolumn{1}{c}{\multirow{4}{*}{0.74}} & \multicolumn{1}{c}{0.78} && \multicolumn{1}{c}{\textit{99.4}} & \multicolumn{1}{c}{\multirow{4}{*}{\textit{0.30}}} & \multicolumn{1}{c}{\textit{0.98}} & & \multicolumn{1}{c}{72.1} & \multicolumn{1}{c}{\multirow{4}{*}{0.88}} & \multicolumn{1}{c}{0.97} & \multicolumn{1}{c}{23.4} & \multicolumn{1}{c}{\multirow{4}{*}{0.40}} & \multicolumn{1}{c}{0.36}\\ 

\multicolumn{1}{c}{} & \multicolumn{1}{c}{} & \multicolumn{1}{c}{LL} & & \multicolumn{1}{c}{94.7} & \multicolumn{1}{c}{} &  \multicolumn{1}{c}{0.98} & \multicolumn{1}{c}{4.9} & \multicolumn{1}{c}{} &  \multicolumn{1}{c}{0.64} && \multicolumn{1}{c}{\textit{98.9}} & \multicolumn{1}{c}{} &  \multicolumn{1}{c}{\textit{0.98}} & & \multicolumn{1}{c}{33.6} & \multicolumn{1}{c}{} &  \multicolumn{1}{c}{0.93} & \multicolumn{1}{c}{0.8} & \multicolumn{1}{c}{} & \multicolumn{1}{c}{0.20}\\ 

\multicolumn{1}{c}{} & \multicolumn{1}{c}{\multirow{2}{*}{\pgdcw}} & \multicolumn{1}{c}{rand} & & \multicolumn{1}{c}{97.2} & \multicolumn{1}{c}{} &  \multicolumn{1}{c}{0.99} & \multicolumn{1}{c}{39.5} & \multicolumn{1}{c}{} &  \multicolumn{1}{c}{0.76} && \multicolumn{1}{c}{\textit{98.6}} & \multicolumn{1}{c}{} &  \multicolumn{1}{c}{\textit{0.90}} & & \multicolumn{1}{c}{68.9} & \multicolumn{1}{c}{} &  \multicolumn{1}{c}{0.97} & \multicolumn{1}{c}{28.4} & \multicolumn{1}{c}{} & \multicolumn{1}{c}{0.33}\\ 

\multicolumn{1}{c}{} & \multicolumn{1}{c}{} & \multicolumn{1}{c}{LL} & & \multicolumn{1}{c}{93.9} & \multicolumn{1}{c}{} &  \multicolumn{1}{c}{0.97} & \multicolumn{1}{c}{4.6} & \multicolumn{1}{c}{} &  \multicolumn{1}{c}{58.3}  && \multicolumn{1}{c}{\textit{93.9}} & \multicolumn{1}{c}{} & \multicolumn{1}{c}{\textit{84.5}} & & \multicolumn{1}{c}{30.4} & \multicolumn{1}{c}{} & \multicolumn{1}{c}{89.1} & \multicolumn{1}{c}{0.8} & \multicolumn{1}{c}{} & \multicolumn{1}{c}{0.17}\\ \midrule

\multicolumn{1}{c}{\multirow{4}{*}{VOC12}} & \multicolumn{1}{c}{\multirow{2}{*}{\pgdx}} & \multicolumn{1}{c}{rand}  & & \multicolumn{1}{c}{99.3} & \multicolumn{1}{c}{\multirow{4}{*}{0.76}} & \multicolumn{1}{c}{0.99} & \multicolumn{1}{c}{54.9} & \multicolumn{1}{c}{\multirow{4}{*}{0.75}} & \multicolumn{1}{c}{0.79} && \multicolumn{1}{c}{99.9} & \multicolumn{1}{c}{\multirow{4}{*}{0.19}} & \multicolumn{1}{c}{0.99} & & \multicolumn{1}{c}{68.7} & \multicolumn{1}{c}{\multirow{4}{*}{0.89}} & \multicolumn{1}{c}{0.97} & \multicolumn{1}{c}{26.4} & \multicolumn{1}{c}{\multirow{4}{*}{0.38}} & \multicolumn{1}{c}{0.36} \\

\multicolumn{1}{c}{} & \multicolumn{1}{c}{} & \multicolumn{1}{c}{LL}  & & \multicolumn{1}{c}{97.5} & \multicolumn{1}{c}{} & \multicolumn{1}{c}{0.98} & \multicolumn{1}{c}{11.3} & \multicolumn{1}{c}{} & \multicolumn{1}{c}{0.61} && \multicolumn{1}{c}{99.8} & \multicolumn{1}{c}{} & \multicolumn{1}{c}{0.99} & & \multicolumn{1}{c}{29.9} & \multicolumn{1}{c}{} & \multicolumn{1}{c}{0.91} & \multicolumn{1}{c}{0.9} & \multicolumn{1}{c}{} & \multicolumn{1}{c}{0.21}\\

\multicolumn{1}{c}{} & \multicolumn{1}{c}{\multirow{2}{*}{\pgdcw}} & \multicolumn{1}{c}{rand} & & \multicolumn{1}{c}{98.5} & \multicolumn{1}{c}{} & \multicolumn{1}{c}{0.99} & \multicolumn{1}{c}{50.3} & \multicolumn{1}{c}{} & \multicolumn{1}{c}{0.76} && \multicolumn{1}{c}{99.7} & \multicolumn{1}{c}{} & \multicolumn{1}{c}{0.95} & & \multicolumn{1}{c}{63.7} & \multicolumn{1}{c}{} & \multicolumn{1}{c}{0.96} & \multicolumn{1}{c}{27.0} & \multicolumn{1}{c}{} & \multicolumn{1}{c}{0.31}\\ 
 
\multicolumn{1}{c}{} & \multicolumn{1}{c}{} & \multicolumn{1}{c}{LL}& & \multicolumn{1}{c}{96.0} & \multicolumn{1}{c}{} & \multicolumn{1}{c}{0.98} & \multicolumn{1}{c}{9.9} & \multicolumn{1}{c}{} & \multicolumn{1}{c}{0.55} && \multicolumn{1}{c}{98.4} & \multicolumn{1}{c}{} & \multicolumn{1}{c}{0.92} & & \multicolumn{1}{c}{26.0} & \multicolumn{1}{c}{} & \multicolumn{1}{c}{0.89} & \multicolumn{1}{c}{1.2} & \multicolumn{1}{c}{} & \multicolumn{1}{c}{0.16}\\ \midrule

\multicolumn{1}{c}{\multirow{4}{*}{\begin{tabular}[c]{@{}c@{}}Internet \\ Photographs\end{tabular}}} & \multicolumn{1}{c}{\multirow{2}{*}{\pgdx}} & \multicolumn{1}{c}{rand} & & \multicolumn{1}{c}{95.4} & \multicolumn{1}{c}{\multirow{4}{*}{0.74}} & \multicolumn{1}{c}{0.99} & \multicolumn{1}{c}{46.3} & \multicolumn{1}{c}{\multirow{4}{*}{0.74}} & \multicolumn{1}{c}{0.80} && \multicolumn{1}{c}{100.0} & \multicolumn{1}{c}{\multirow{4}{*}{0.17}} & \multicolumn{1}{c}{0.99} & & \multicolumn{1}{c}{58.4} & \multicolumn{1}{c}{\multirow{4}{*}{0.89}} & \multicolumn{1}{c}{0.97} & \multicolumn{1}{c}{19.8} & \multicolumn{1}{c}{\multirow{4}{*}{0.43}} & \multicolumn{1}{c}{0.35}\\ 

\multicolumn{1}{c}{} & \multicolumn{1}{c}{} & \multicolumn{1}{c}{LL} & & \multicolumn{1}{c}{92.1} & \multicolumn{1}{c}{} & \multicolumn{1}{c}{0.98} & \multicolumn{1}{c}{12.2} & \multicolumn{1}{c}{} & \multicolumn{1}{c}{0.34} && \multicolumn{1}{c}{99.6} & \multicolumn{1}{c}{} & \multicolumn{1}{c}{0.98} & & \multicolumn{1}{c}{26.8} & \multicolumn{1}{c}{} & \multicolumn{1}{c}{0.93} & \multicolumn{1}{c}{0.5} & \multicolumn{1}{c}{} & \multicolumn{1}{c}{0.24}\\ 

\multicolumn{1}{c}{} & \multicolumn{1}{c}{\multirow{2}{*}{\pgdcw}} & \multicolumn{1}{c}{rand} & & \multicolumn{1}{c}{95.4} & \multicolumn{1}{c}{} & \multicolumn{1}{c}{0.98} & \multicolumn{1}{c}{47.5} & \multicolumn{1}{c}{} & \multicolumn{1}{c}{0.77} && \multicolumn{1}{c}{99.8} & \multicolumn{1}{c}{} & \multicolumn{1}{c}{0.96} & & \multicolumn{1}{c}{55.5} & \multicolumn{1}{c}{} & \multicolumn{1}{c}{0.95} & \multicolumn{1}{c}{22.4} & \multicolumn{1}{c}{} & \multicolumn{1}{c}{0.33}\\ 

\multicolumn{1}{c}{} & \multicolumn{1}{c}{} & \multicolumn{1}{c}{LL} & & \multicolumn{1}{c}{91.2} & \multicolumn{1}{c}{} & \multicolumn{1}{c}{0.96} & \multicolumn{1}{c}{11.4} & \multicolumn{1}{c}{} & \multicolumn{1}{c}{0.62} && \multicolumn{1}{c}{99.0} & \multicolumn{1}{c}{} & \multicolumn{1}{c}{0.93} & & \multicolumn{1}{c}{23.3} & \multicolumn{1}{c}{} & \multicolumn{1}{c}{0.91} & \multicolumn{1}{c}{0.6} & \multicolumn{1}{c}{} & \multicolumn{1}{c}{0.19}\\ \midrule

\multicolumn{1}{c}{\multirow{4}{*}{\begin{tabular}[c]{@{}c@{}}Gaussian \\ Noise\end{tabular}}} & \multicolumn{1}{c}{\multirow{2}{*}{\pgdx}} & \multicolumn{1}{c}{rand} & & \multicolumn{1}{c}{\textbf{100}} & \multicolumn{1}{c}{\multirow{4}{*}{0.92}} & \multicolumn{1}{c}{1.00} & \multicolumn{1}{c}{\textbf{87.9}} & \multicolumn{1}{c}{\multirow{4}{*}{0.52}} & \multicolumn{1}{c}{0.86} && \multicolumn{1}{c}{48.5} & \multicolumn{1}{c}{\multirow{4}{*}{0.41}} & \multicolumn{1}{c}{0.45} & & \multicolumn{1}{c}{\textbf{79.0}} & \multicolumn{1}{c}{\multirow{4}{*}{0.91}} & \multicolumn{1}{c}{0.99} & \multicolumn{1}{c}{\textbf{29.1}} & \multicolumn{1}{c}{\multirow{4}{*}{0.31}} & \multicolumn{1}{c}{0.32}\\ 

\multicolumn{1}{c}{} & \multicolumn{1}{c}{} & \multicolumn{1}{c}{LL} && \multicolumn{1}{c}{100} & \multicolumn{1}{c}{} & \multicolumn{1}{c}{1.00} & \multicolumn{1}{c}{7.1} & \multicolumn{1}{c}{} & \multicolumn{1}{c}{0.39} && \multicolumn{1}{c}{0} & \multicolumn{1}{c}{} & \multicolumn{1}{c}{0} && \multicolumn{1}{c}{0} & \multicolumn{1}{c}{} & \multicolumn{1}{c}{0} & \multicolumn{1}{c}{0} & \multicolumn{1}{c}{} & \multicolumn{1}{c}{0}\\ 

\multicolumn{1}{c}{} & \multicolumn{1}{c}{\multirow{2}{*}{\pgdcw}} & \multicolumn{1}{c}{rand} && \multicolumn{1}{c}{100.0} & \multicolumn{1}{c}{} & \multicolumn{1}{c}{1.00} & \multicolumn{1}{c}{87.3} & \multicolumn{1}{c}{} & \multicolumn{1}{c}{0.81} && \multicolumn{1}{c}{47.7} & \multicolumn{1}{c}{} & \multicolumn{1}{c}{0.29} && \multicolumn{1}{c}{79.7} & \multicolumn{1}{c}{} & \multicolumn{1}{c}{0.97} & \multicolumn{1}{c}{28.4} & \multicolumn{1}{c}{} & \multicolumn{1}{c}{0.30}\\ 

\multicolumn{1}{c}{} & \multicolumn{1}{c}{} & \multicolumn{1}{c}{LL} && \multicolumn{1}{c}{100} & \multicolumn{1}{c}{} & \multicolumn{1}{c}{1.00} & \multicolumn{1}{c}{4.9} & \multicolumn{1}{c}{} & \multicolumn{1}{c}{0.31} && \multicolumn{1}{c}{0} & \multicolumn{1}{c}{} & \multicolumn{1}{c}{0} && \multicolumn{1}{c}{0} & \multicolumn{1}{c}{} & \multicolumn{1}{c}{0} & \multicolumn{1}{c}{0} & \multicolumn{1}{c}{} & \multicolumn{1}{c}{0}\\
\bottomrule
\end{tabular}
}
\vspace{-10pt}
\end{table*}

\subsubsection{Baseline models are highly vulnerable to \oodAdvExamples}\label{subsubsec: baseline_attack}
Fig. \ref{fig:cifar_clean_pgd} shows the target success rate with adversarial examples generated from different OOD datasets for the WRN-28-10 network (Table~\ref{table:dataset_and_model_details}) trained on the CIFAR-10 dataset. The x-axis represents the maximum $l_\infty$ perturbation for the \textsf{PGD-xent} attack used to generate the adversarial examples. The results show that similar to adversarial examples generated from in-distribution images, \emph{\oodAdvExamples also achieve a high target success rate}. For example, the target success rate increases rapidly to 100$\%$ for both in- and \oodlong data.
 

\subsubsection{\oodattacks on adversarially trained models} \label{subsubsec: ood_adv_trained}

We now evaluate the robustness of adversarially trained models (iterative adversarial training~\cite{madry_towards_2017}) to \oodAdvExamples. Previous work \cite{goodfellow2014explaining,tramer2017ensemble,madry_towards_2017, kannan2018adversarial} has shown that adversarial training can significantly increase the model robustness against adversarial examples generated from in-distribution data. For example, for the WRN-28-10 network trained on CIFAR-10, adversarial training reduces the target success rate from 100$\%$ to 22.9$\%$ for the \textsf{PGD-xent} attack with $\epsilon$ equal to 8,
as shown in Figure \ref{fig:cifar_adv_trained_pgd}. 
%

\noindent \textbf{Experimental details:} For the MNIST and CIFAR-10 datasets, we use the iterative adversarial training approach proposed in Madry et al. \cite{madry_towards_2017}.  Models corresponding to MNIST, CIFAR-10 datasets in this experiment are $M_1$, WRN-28-10-A (Table~\ref{table:dataset_and_model_details}) respectively. Each model is adversarially trained with an $L_\infty$ perturbation budget ($\epsilon$) equal to 0.3, 8 respectively. \par 

\noindent \textbf{\oodAdvExamples generated from OOD datasets (except MNIST) achieve high target success rates for multiple adversarial trained models and datasets:} Fig.~\ref{fig:cifar_adv_trained_pgd} shows target success rate of \oodAdvExamples for a WRN-28-10 network, trained on the CIFAR-10 dataset using adversarial training, with different perturbation budget ($\epsilon$) for \textsf{PGD-xent} attack. It shows that though adversarial training improves robustness for in-distribution dataset (CIFAR-10), \oodAdvExamples can achieve up to $4\times$ higher target success rate compared adversarial examples generated from in-distribution images. \par 
Table~\ref{table:default_robust_defenses} presents the detailed results for different datasets, attacks, and label selection. We can see the improvement in target success rate with \oodAdvExamples. When using the VOC12 dataset to generate \oodAdvExamples, we can achieve around $66.7\times$ and $2.4\times$ improvement in target success rate compared to in-distribution attacks for MNIST and CIFAR-10 models respectively. The mean classification confidence for \oodAdvExamples is also competitive with adversarial examples generated from in-distribution data and typically higher than unmodified OOD data. However, achieving the least likely target label (LL) is a significantly harder objective for evasion attacks with both in- and \oodlong data. For each dataset, we observe that \textsf{PGD-CW} is relatively less successful than \textsf{PGD-xent} attack.
%
%

\subsubsection{OOD attacks on robust training with the convex polytope relaxation} \label{subsubsec: ood_prov_train}

Robust training using convex polytope relaxation from Wong et al.~\cite{kolter2017provable, wong2018scaling} provides a \emph{provable} upper bound on the adversarial error and thus on target success rate. \par 

\noindent \textbf{Experiment Details:} We use the models from Wong et al.~\cite{kolter2017provable, wong2018scaling} for MNIST and CIFAR-10 dataset. The corresponding Models are $M_2$ and WRN-28-1 respectively (Table~\ref{table:dataset_and_model_details}). These models achieve the state-of-the-art minimum provable adversarial error with $L_\infty$ perturbation ($\epsilon$) equals to 0.1 and 2 for MNIST and CIFAR-10 data respectively. With more complex dataset such as CIFAR-10, this defense is limited to the use of a small perturbation budget in training to avoid reducing the performance on unmodified inputs. For example, training the WRN-28-1 network using convex polytope relaxation on CIFAR-10 dataset with a more realistic $\epsilon$ of 8 achieves only 27.1$\%$ classification accuracy on unmodified CIFAR-10 test data.
Given that the defense approach cannot simultaneously maintain high benign accuracy for CIFAR-10 while using a more realistic $\epsilon$ equals to 8, we make a design choice of $\epsilon$ equals to 2 in training the CIFAR-10 model and $\epsilon$ equals to 8 for adversarial example generation. On the other hand, for simpler dataset such as MNIST, we continue to use the same $\epsilon$ equals to 0.1 for both training and adversarial example generation.

%

\noindent \textbf{Robust training with convex polytope relaxation lacks robustness to \oodAdvExamples:} Table~\ref{table:default_robust_defenses} shows the experimental results for MNIST and CIFAR10 models using the provable defense approach of convex polytope relaxation~\cite{kolter2017provable, wong2018scaling}. Though this approach significantly improves the robustness for in-distribution adversarial examples, it lacks robustness to \oodAdvExamples. For the model trained on MNIST, the target success rate increases from 0.6$\%$ to 72.1$\%$ by using ImageNet as a source of \ood data with $\epsilon$ = 0.1 and $\pgdx$ attack. Similar success rate is observed with other OOD datasets for this model.
For the model trained on CIFAR-10 dataset, the target success rate increases from 15.1$\%$ to 29.1$\%$ with the use of \oodAdvExamples generated from Gaussian noise. The relatively poor performance of adversarial examples for the CIFAR-10 model could be due to the \emph{poor classification accuracy of this model}, where it achieves only 66.2$\%$ classification accuracy on the CIFAR-10 images.
We argue that the principle behind this defense is not robust to \oodAdvExamples demonstrated by OOD attacks on the provably trained MNIST model.



\subsubsection{Discussion: Impact of OOD dataset} \label{subsubsec: Discussion_adv_defense}
In this subsection, we further discuss the influence of dataset selection and robust learning on the success of evasion attacks. We observe in Table \ref{table:default_robust_defenses} that the target success rate is affected by the choice of the OOD dataset. In particular, we observe that the target success rate for MNIST dataset is significantly lower than both in-distribution and other OOD datasets. We speculate that this behavior could arise due to the specific semantic structure of MNIST images. Nevertheless, we emphasize that the threat posed by OOD adversarial examples still persists, since \emph{adversarial examples from multiple other OOD datasets achieve high target success rates}. 

\subsection{Evading Adversarial Example Detectors and Secondary Defenses}\label{subsec: adv_detectors}
\subsubsection{Adversarial detectors.} 
Previous work \cite{he2017ensembleweak,carlini2017magnet} has shown that both adversarial detectors based on Feature Squeezing~\cite{xu2017feature} or MagNet~\cite{meng2017magnet} approach can be evaded with adaptive white-box attacks accounting for the detector mechanism to generate adversarial examples from in-distribution data. Our results show that similar to in-distribution data, \emph{these adversarial detectors don't provide robustness to adversarial examples generated from OOD data}. For feature squeezing, we also show that OOD data requires a smaller $L_2$ perturbation budget than in-distribution data for a similar target success rate of corresponding adversarial examples. We further show that \oodAdvExamples can achieve up to $97.3\%$ target success rate in presence of MagNet on a model trained on the CIFAR-10 dataset. We provide detailed results in Appendix~\ref{appsec:adv_detectors}.\par 
\subsubsection{Adversarial logit pairing.}
The robust training based on iterative adversarial training by Madry et al.~\cite{madry_towards_2017} works well for small-scale datasets but does not provide robustness with Imagenet. Adversarial logit pairing \cite{kannan2018adversarial} extends this approach to provide robustness on Imagenet. However, ALP suffers from loss of robustness for in-distribution adversarial examples when the number of attack iterations is increased \cite{breakingALP2018}. We show that this vulnerability also exists for \oodAdvExamples (Table~\ref{table:default_robust_defenses}). We use ResNet-v2-50 (Table~\ref{table:dataset_and_model_details}) and perturbation budget ($\epsilon$) of 16.

\subsection{Towards real-world attacks} \label{subsec: real-world}
In this section, we demonstrate how \ood adversarial examples may affect real-world ML systems.

\subsubsection{Attacking Content Moderation systems} \label{subsubsec: content_mod_attack} To achieve low false positive rate (FPR), ML classifiers deployed in the real-world are also expected to detect OOD inputs. The is because a high FPR can significantly affect the performance \cite{metnudity} and cost \cite{ponemon2018study} of the service provided by the these models. For example, the Metropolitan police in London are attempting to use computer vision to detect nudity in photographs, but a high FPR is reportedly occurring due to the prevalence of desert scenes as wallpapers etc. \cite{metnudity}. This example represents an inadvertent \emph{denial-of-service (DoS) attack}, where a large number of false positives affects the effectiveness of the automated content moderation system.

We use \ood adversarial examples to carry out a similar DoS attack on Clarifai's content moderation model \cite{clarifai}, by \emph{classifying clearly unobjectionable content as objectionable with high confidence}. In a real deployment, a deluge of such data will force human content moderators to spend time reviewing safe content. Sybil attacks \cite{yu2006sybilguard,danezis2009sybilinfer} can enable attackers to create a plethora of fake accounts and upload large amounts of \ood adversarial examples.

\begin{figure}[t]
	\centering
	{
		\includegraphics[width=1cm,height=1cm]{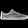}%
		\label{subfig: fmnist_benign}%
	}
	\hspace{10pt}
	{
		\includegraphics[width=1cm,height=1cm]{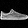}%
		\label{subfig: fmnist_adv}%
	}
	\caption{\ood adversarial examples against Clarifai's Content Moderation model. \textbf{Left}: original image, classified as `safe' with a confidence of 0.96. \textbf{Right}: adversarial example with $\epsilon=16$, classified as `explicit' with a confidence of 0.9.}
	\label{fig: clarifiai_attack}
	\vspace{-15pt}
\end{figure}

\noindent \textbf{Fooling the Clarifai model:} Using the query-based black-box attacks proposed by Bhagoji et al. \cite{bhagoji_eccv_2018}, we construct \ood adversarial examples for Clarifai's content moderation model. It provides confidence scores for the input image belonging the 5 classes `safe', `suggestive', `explicit', `gore' and `drugs', and is accessible through an API. We use 10 images each from the MNIST, Fashion-MNIST \cite{xiao2017fashion} and Gaussian Noise datasets to generate \ood adversarial examples. All of these images are initially classified as `safe' by the model. In Figure \ref{fig: clarifiai_attack}, we show a representative attack example. Our attack is able to successfully generate \ood adversarial examples for the 4 classes apart from `safe' for all 30 images with 3000 queries on average and a mean target confidence of 0.7.
\subsubsection{Physically-realizable attacks on traffic signs}  \label{subsubsec: traffic_sign_attack}
\begin{figure}[t]
	\centering
	\includegraphics[width=0.47\textwidth]{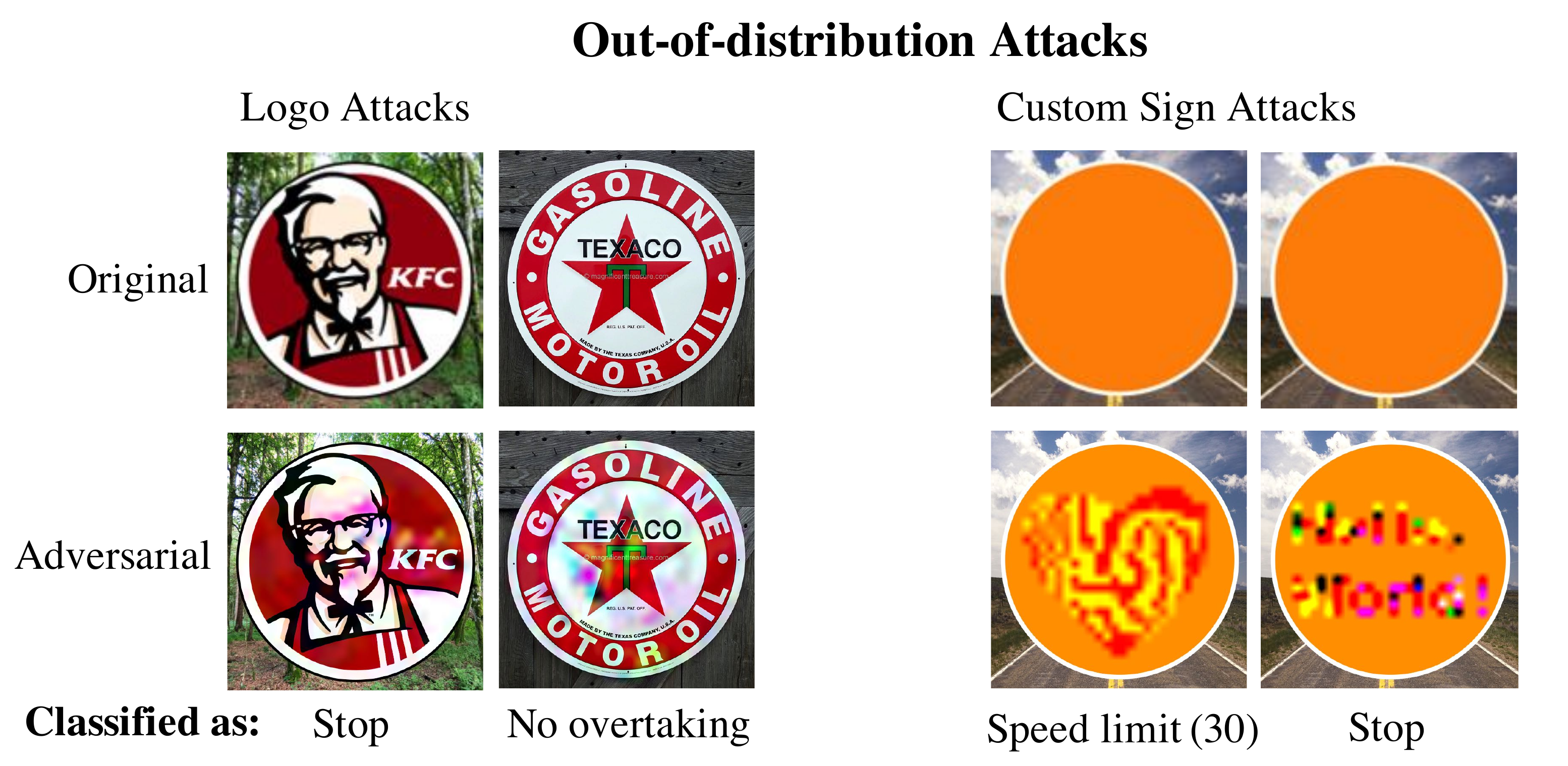}
	\caption{\ood adversarial examples for a traffic sign recognition pipeline. These adversarial examples are \emph{classified as the desired target traffic sign with high confidence} under a variety of physical conditions when printed out.}
	\label{fig: traffic_adv_ood}
	\vspace{-15pt}
\end{figure}

Traffic sign recognition systems are intended for operation in the real-world, where they encounter arbitrary objects in the environment. These objects can be modified to become physical \ood adversarial examples which are detected and classified with high confidence as traffic signs. We demonstrate attack success with both imperceptible perturbations in a \ood logo attack and unconstrained perturbations within a mask in a custom sign attack. 
\ood adversarial examples in Figure \ref{fig: traffic_adv_ood} are detected and classified with high confidence as traffic signs (by a CNN with 98.5\% accuracy on test data) over a range of physical conditions when printed out. The targeted attack success rate is 95.2\% for the custom sign attack. Details of these attacks and further results are in our short workshop paper \cite{Sitawarin17}. 

\section{Towards Robust Open-world Deep Learning}\label{sec:defense}
In this section, we present experimental results for our proposed hybrid combination of iterative adversarial training and selective prediction to enhance classifier robustness against \oodAdvExamples. We experiment with image classification task with CIFAR-10 dataset as in-distribution and multiple other datasets as the source of \oodlong images. \par 
\noindent \textbf{Setup:} In this experiment we use WRN-28-10 model with the hyper-parameters from Madry et al.~\cite{madry_towards_2017} for iterative adversarial training and CIFAR-10 as the in-distribution dataset (Table II). To train a robust classifier with a background class, we use 5,000 images from one of the MNIST, ImageNet, VOC12, and Internet Photographs datasets. An image will be specified as OOD if it is classified to the background class. We select random target labels from CIFAR-10 classes as the desired target class (excluding the predicted class for the unmodified input), and compute the target success rate. This metric allows us to capture the adversary's success at both evading detection and achieving targeted misclassification with \oodAdvExamples. The number of test images and perturbation budget ($\epsilon$) is 1000 and 8 respectively.\par 
\begin{figure}[!htb]
    \centering
    \includegraphics[width=0.8\linewidth]{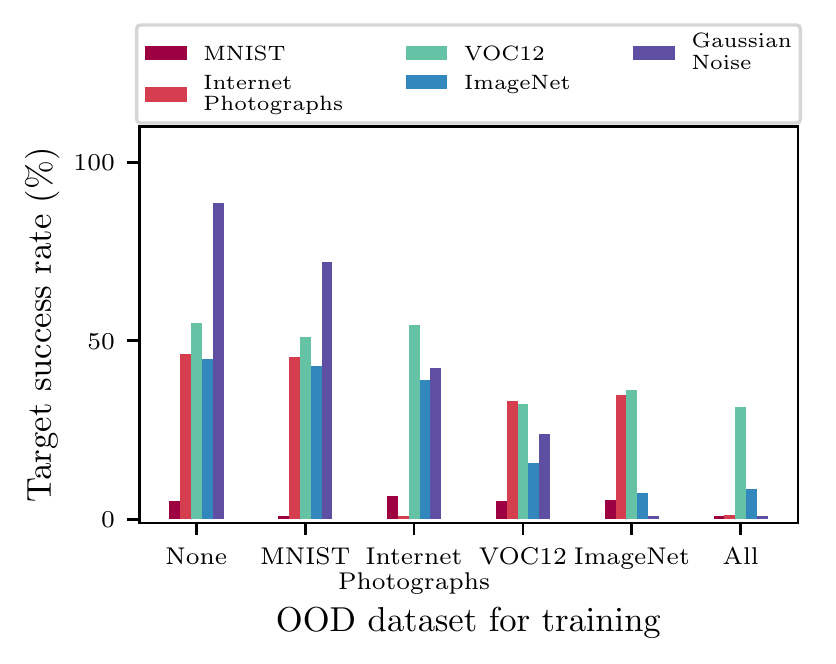}
     \vspace{-5pt}
    \caption{Target success rate of \oodAdvExamples (lower is better). The classifier is adversarially trained on in-distribution inputs along with a small subset of OOD data. It shows that datasets such as VOC12 and ImageNet provide a high inter-dataset and intra-dataset generalization for detection of adversarial OOD inputs.}
   \label{fig:background_class_adv_scs}
   \vspace{-5pt}
\end{figure}

\noindent \textbf{Small subset of OOD data can enable robust detection.} With each of the OOD dataset, we observe that using only 5,000 training images leads to significant decrease in the success of adversarial attacks from these datasets (Fig.~\ref{fig:background_class_adv_scs}). For example, after adversarial training with only 5,000 training images (out of 1.2 million) from ImageNet, the target success of \oodAdvExamples from ImageNet dataset decreases from 44.9\% to 7.4\%. \par 
\noindent \textbf{Robust training with one \ood dataset can generalize to multiple other OOD datasets.} From Figure \ref{fig:background_class_adv_scs}, we observe that adversarial training with one dataset also decreases attack success of OOD adversarial examples from other datasets. This effect is significant for feature-rich datasets such as ImageNet and VOC12. For example, using VOC12 for training reduces the target success rate of \oodAdvExamples from ImageNet from 44.9\% to 15.8\%. \par 
\noindent \textbf{Multiple OOD datasets can be combined for robust detection:} By including 5,000 images from each of the four OOD datasets in robust training, we demonstrate that a single network can learn robust detection for each of them. We observe that the combination of all datasets is constructive as the best results are achieved when multiple datasets are used in training. \par 
\noindent \textbf{Small impact on in-distribution performance:} Robust training with OOD data has a small impact on classifier performance for in-distribution data. The maximum decrease in benign classification accuracy is 1\% (for single OOD dataset) and 3.1\% (for all four OOD datasets). The robust accuracy remains largely unchanged. Detailed results are in Table \ref{table:background_class} in Appendix~\ref{appsec:ood_defense}. \par 
\noindent \textbf{Discussion and limitations.}
Our results highlight that it is feasible to robustly classify one or multiple OOD datasets along with in-distribution data using a semi-supervised learning approach. However, the key challenge in this domain is to achieve robustness against all \ood inputs. As a first step, our approach and results motivate the design of robust unsupervised OOD detectors for deep learning. They also highlight the need for rigorous evaluation methods to determine the robustness of an open-world learning system against all possible adversarial examples.

\section{Conclusion}\label{sec:discussion}
In this paper, we investigated evasion attacks in the open-world learning model and defined \ood adversarial examples, which represent a new attack vector on ML models used in practice. We found that existing \ood detectors are insufficient to deal with this threat. Further, assumptions regarding the source of adversarial examples, namely, in-distribution data, have led to tailored defenses. We showed that these state-of-the-art defenses exhibit increased vulnerability to \ood adversarial examples, which makes their deployment challenging. With these findings in mind, we took a first step at countering \ood adversarial examples using adversarial training with background class augmented classifiers. We now urge the community to consider the exploration of strong defenses against open-world evasion attacks.


\bibliographystyle{ACM-Reference-Format}
\bibliography{OOD}

\appendix
\section{Additional details for the datasets} \label{sec:appendix_dataset_details}
Table \ref{table:dataset_appendix} provides the details of the datasets used in this work. In particular, it provides the state-of-the-art top-1 classification accuracy for each dataset, which can be used to compare with the performance of models used in the paper.
\begin{table}[!htb]
\caption{Details of the three dataset used as as source of in-distribution images in this work.}
\label{table:dataset_appendix}
\resizebox{0.48\textwidth}{!}{%
\begin{tabular}{ccccc}
\toprule
 & \begin{tabular}[c]{@{}c@{}}Number of\\ classes\end{tabular} & \multicolumn{2}{c}{\begin{tabular}[c]{@{}c@{}}Number of \\ images (in thousands)\end{tabular}} & \begin{tabular}[c]{@{}c@{}}State-of-the-art \\  classification \\ accuracy (\%)\end{tabular} \\ \cline{3-4}
 &  & Train & Test/Validation &  \\ \midrule 
MNIST & 10 & 60 & 10 & 99.79 \cite{mnistsota} \\
CIFAR-10 & 10 & 50 & 10 & 97.69 \cite{yamada2018shakedrop} \\
\begin{tabular}[c]{@{}c@{}}ImageNet\\ (ILSVRC12)\end{tabular} & 1000 & 1281 & 50 & 82.7~\cite{imagnetsota} \\
\bottomrule
\end{tabular}
}
\end{table}

We also use VOC12 dataset as a source of OOD images. In addition, the two datasets we constructed as sources of \ood data are detailed below:
\begin{enumerate}
	\item \textbf{Internet photographs}: We sample $10000$ random natural images from the Picsum random image generator \cite{picsum}, which is a source of \emph{real-world} photographs. These images have dimensions of between 300$\times$300 to 800$\times$800 pixels.  
	
	\item \textbf{Gaussian noise}:  We synthesize 10000 random images using a Gaussian distribution for each pixel value. We set the mean ($\mu$) of the Gaussian distribution equal to $127$ and the standard deviation ($\sigma$) to 50. We don't select a very low or high standard deviation as it leads to very sharp or uniform distribution in the valid pixel range ([0, 255]) respectively.
\end{enumerate}

\begin{table}[t]
	\fontsize{30pt}{30pt}\selectfont
	\renewcommand{\arraystretch}{1.5}
	\caption{Minimum and maximum of expected confidence over all output classes of a WRN-28-10 model (trained on CIFAR-10 dataset) for unmodified and adversarial OOD inputs. High $Min.$ and $Max.$ values \emph{demonstrate that \oodAdvExamples can achieve target classification to any output class with high confidence}.
	}
	\label{table:cifar_confidence}
	\resizebox{0.48\textwidth}{!}{%
		\begin{tabular}{ccccccc}
			\toprule
			& & \multicolumn{5}{c}{OOD Dataset} \\ \cline{3-7} 
			\begin{tabular}[c]{@{}c@{}}Image \\ type\end{tabular} & metric & MNIST & ImageNet & VOC12 & \begin{tabular}[c]{@{}c@{}}Random\\ photographs\end{tabular} & \begin{tabular}[c]{@{}c@{}}Gaussian \\ noise\end{tabular} \\ \midrule
			\multirow{2}{*}{\begin{tabular}[c]{@{}c@{}}Unmodified\end{tabular}} & $Min.$ & 0.00 & 0.01 & 0.02 & 0.00 & 0.00 \\
			& $Max.$ & 0.62 & 0.29 & 0.17 & 0.20 & 0.97 \\ \midrule
			\multirow{2}{*}{Adversarial} & $Min.$ & 0.99 & 0.99 & 0.99 & 0.99 & 0.99 \\
			& $Max.$ & 1.00 & 1.00 & 1.00 & 1.00 & 1.00 \\ \bottomrule
		\end{tabular}
	}
\vspace{-10pt}
\end{table}

\section{Comparing the impact of unmodified and adversarial OOD data} \label{appsec: benign_vs_adv}
Since the classifier $f$ has never encountered inputs $\bfx$ drawn from $P^{\text{out}}_{X}$, it is unclear what its behavior will be. An empirical analysis of the behavior of state-of-the-art classifiers on \emph{unmodified} OOD data motivates the need, from an adversarial perspective, to generate \oodAdvExamples. 

\noindent \textbf{Methodology:} We use a wide residual net (WRN-28-10) model trained on CIFAR-10 dataset to illustrate the differences between unmodified and adversarial \ood data. The adversary first selects a target class $t \in \mathcal{Y}$ and then samples 1,000 images from OOD datasets which should be classified as class $t$ ($\mathbf{X}^{\text{out}}_t$). The adversarial counterparts of these images are generated using the PGD attack with $L_{\infty}$ norm constraints \cite{madry_towards_2017}. Further details of attacks, models and datasets are in Section \ref{sec:setup}. For each set of examples $\mathbf{X}^{\text{out}}_t$ with target $t$, $\mathbb{E}_{\bfx \sim \mathbf{X}_t^{\text{out}}} [g(\bfx)(t)]$ is the expected confidence. We then report the minimum $Min.$ and maximum $Max.$ of this expectation over targets $t$ in Table~\ref{table:cifar_confidence}.

\noindent \textbf{Unmodified \ood data:} For unmodified OOD images, the adversary doesn't have control over target classification and output confidence. As an example, for unmodified MNIST images as OOD inputs, one class is rarely predicted (resulting in $Min.$ = 0), while another class is predicted with an average confidence of 0.62. This indicates that while some targets can be met confidently with unmodified \ood data, not all outputs are equally likely for a given $P_{X}^{\text{out}}$ and that targets $T \in \mathcal{Y}$ may not be reachable with unmodified OOD data.

\noindent \textbf{High-confidence targeted misclassification}: On the other hand, it is clear that for \ood adversarial examples, any desired target class can be achieved with high confidence as both the $Min.$ and $Max.$ values are high. This holds across \ood datasets.

\section{Detailed Results: Adversarial Example Detectors}\label{appsec:adv_detectors}
In this section, we evaluate OOD attacks against the state-of-art adversarial example detectors, including feature squeezing~\cite{xu2017feature} and MagNet~\cite{meng2017magnet} (recall Section \ref{subsubsec: back_adv_detection}). Target success rate in presence of these detectors refers to the percentage of adversarial examples which both evade the detectors and achieve target label after classification. \par 
\noindent \textbf{Summary of results.} Previous work \cite{he2017ensembleweak,carlini2017magnet} has shown that these adversarial detectors can be evaded with adaptive adversaries by successfully generating adversarial examples from in-distribution data. Our results show that similar to in-distribution data, \emph{these adversarial detectors don't provide robustness to adversarial examples generated from OOD data}. For feature squeezing, we also show that OOD data requires a smaller $L_2$ perturbation budget than in-distribution data for a similar target success rate of corresponding adversarial examples. We further show that \oodAdvExamples can achieve up to $97.3\%$ target success rate in presence of MagNet on a model trained on the CIFAR-10 dataset.
\begin{table}[t]
\fontsize{15pt}{15pt}\selectfont
\renewcommand{\arraystretch}{1.3} 
\caption{Target success rate for adversarial examples with random target labels from different datasets in presence of adversarial detectors, including feature squeezing~\cite{xu2017feature} and MagNet~\cite{meng2017magnet}. Similar to adversarial examples generated from in-distribution inputs, \oodAdvExamples are also able to evade the adversarial detectors with a high success rate (with the exception of MagNet for models trained on the MNIST dataset, which is explained in Section~\ref{section:magnet}). 
}
\label{table:adv_detectors_short}
\resizebox{\linewidth}{!}{%
\begin{tabular}{cccccccc}
\toprule
\multicolumn{1}{c}{} && \multicolumn{3}{c}{Feature squeezing~\cite{xu2017feature}} && \multicolumn{2}{c}{\begin{tabular}[c]{@{}c@{}} MagNet \cite{meng2017magnet}\end{tabular}} \\ \cline{3-5} \cline{7-8}
\multicolumn{1}{c}{\begin{tabular}[c]{@{}c@{}}Test ($\downarrow$)~\textbackslash{}~Train ($\rightarrow$)\\ dataset\end{tabular}} && \multicolumn{1}{c}{MNIST} & \multicolumn{1}{c}{CIFAR-10} & \multicolumn{1}{c}{ImageNet} && \multicolumn{1}{c}{\begin{tabular}[c]{@{}c@{}}MNIST \\ ($\epsilon = 0.3$) \end{tabular}} & \multicolumn{1}{c}{\begin{tabular}[c]{@{}c@{}} CIFAR-10 \\ ($\epsilon = 8$) \end{tabular}} \\ \midrule
MNIST && 98.1 & 100.0 & 3.12  && \textbf{32.3} & 18.0 \\
CIFAR-10 && 99.2 & 100.0 & 96.1  && 0.7 & 90.1 \\
ImageNet && 99.3 & 100.0 & 85.1  && 0.8 & 92.5\\
VOC12 && 99.3 & \textbf{100.0} & 96.0 && 0.8 & 96.9 \\
\begin{tabular}[c]{@{}c@{}} Internet \\ Photographs\end{tabular} && 98.1 & 100.0 & 85.1 && 1.2 & \textbf{97.3} \\
\begin{tabular}[c]{@{}c@{}} Gaussian  \\ Noise \end{tabular} && \textbf{100.0} & 100.0 & 25.0 && 0.0 & 93.9 \\ \bottomrule
\end{tabular}
}
\end{table}


\bcomment{
As shown in Table \ref{table:inputpreprocessing}, with the increase of $\epsilon$, the target success rate also increases up to 100\%. Note that during the experiments, we set the threshold values of detectors such that 5\% of benign images will be mis-detected as adversarial (the original paper \cite{meng2017magnet} to reject 1\% or 0.5\% benign images), by doing this, we actually increase the difficulty of the adversary to evade the detector. Different from Section V where OOD examples will have higher success rate than in-distribution adversarial examples, here in-distribution adversarial images (CIFAR10) achieve similar attack performance as OOD dataset. That is because MagNet cannot defend against an adaptive adversary.

\textbf{takeaway: our proposed OOD adversarial examples can mislead MagNet successfully, resulting in similar target success rate as in-distribution white-box adversarial attacks.}

\noindent \textbf{Takeaway: In summary, this section highlights that adding an input pre-processing steps doesn't increase the robustness to \oodAdvExamples~(given a white-box threat model).
The importance of these experiments is that the input pre-processing steps behave differently for In and out distribution input. For eg., Defense-GAN adopts projection on training data manifold as input pre-processing. An adaptive attack for In-distribution inputs succeeds by assuming that the projected output is likely to close the given input. Such an assumption is not valid and may require some further tweaking in attack method.}

}

\subsection{Feature Squeezing}
\noindent \textbf{Experimental details.} We use the joint squeezers (bit depth reduction, non-local smoothing, and median filtering) recommended by the Xu et al.~\cite{xu2017feature} for MNIST, CIFAR-10, and ImageNet dataset. The corresponding models for these datasets are $M_3$, WRN-28-10, and MobileNet (see Table~\ref{table:dataset_and_model_details}). We use Adam~\cite{kingma2014adam} solver with $L_2$ perturbation and straight-through-estimator approach from Athalye et al.~\cite{athalye2018obfuscated}. We report the target success rate of adversarial examples for feature squeezing in Table~\ref{table:adv_detectors_short}. Based on these results, the following two conclusions can be drawn: \par 
\noindent \textbf{Adversarial examples generated from both in- and out-of- distribution images achieve high success rate}:
For all target models, we observe that the adaptive adversary can achieve a high success rate (in most cases, around $100\%$) for OOD and in-distribution adversarial examples. This is because Feature Squeezing uses a non-adversarial trained model, which lacks robustness once the adversary can calculate information gradients in presence of the squeezers. \par 
\noindent \textbf{Most \oodAdvExamples from require less input perturbation}:
In addition to achieving high target success rate, adversarial examples generated from OOD datasets, also require less input perturbation. For networks trained on MNIST dataset, the mean $l_2$ norm of input perturbations for MNIST and CIFAR-10 datasets is 5.67 and 2.12 respectively. Similarly, for CIFAR-10 trained model, the mean perturbation required for CIFAR-10 and ImageNet images is 1.24 and 1.00 respectively.
\subsection{MagNet} \label{section:magnet}
%
We follow the code released by Meng and Chen~\cite{meng2017magnet} to train the MNIST and CIFAR-10 classifiers and corresponding autoencoders\footnote{\url{https://github.com/Trevillie/MagNet}}. 
The models for MNIST and CIFAR-10 are $M_3$ and All Convolution Net \cite{springenberg2014striving}, as shown in Table \ref{table:dataset_and_model_details}. We consider an adaptive adversary which incorporates the distance between input and projected output from the autoencoder into the loss function. 
The summary of key results, presented in Tabel~\ref{table:adv_detectors_short}, for MagNet is:\par
%
\noindent \textbf{OOD adversarial examples can achieve high target success rate for model trained on CIFAR-10.} As shown in Table \ref{table:adv_detectors_short} in the setting of \textsf{PGD-xent} attacks and random target labeling, \oodAdvExamples except MNIST achieve a target success rate higher than $92\%$ in presence of MagNet for the model trained on CIFAR-10 dataset. The target success rate of adversarial examples generated from in-distribution data is $90.1\%$.\par
\noindent \textbf{\oodAdvExamples have reduced effectiveness for MNIST models.}
For the model trained on the MNIST dataset, we observe that the distance values between the projection of autoencoder and original OOD input tend to be around 10 times the threshold used by MagNet. This makes it hard, even for the adaptive adversary, to reduce the distance while achieving the target classification at the output for \oodAdvExamples. We observe that additional pre-processing techniques, such as moving pixels of input images close to zero i.e., similar to the background of most images in MNIST, can increase the target success rate from 0 to up to 12\%.

\section{Additional results: Robust open-world machine learning} \label{appsec:ood_defense}
\begin{figure}[!htb]
    \centering
    \includegraphics[width=\linewidth]{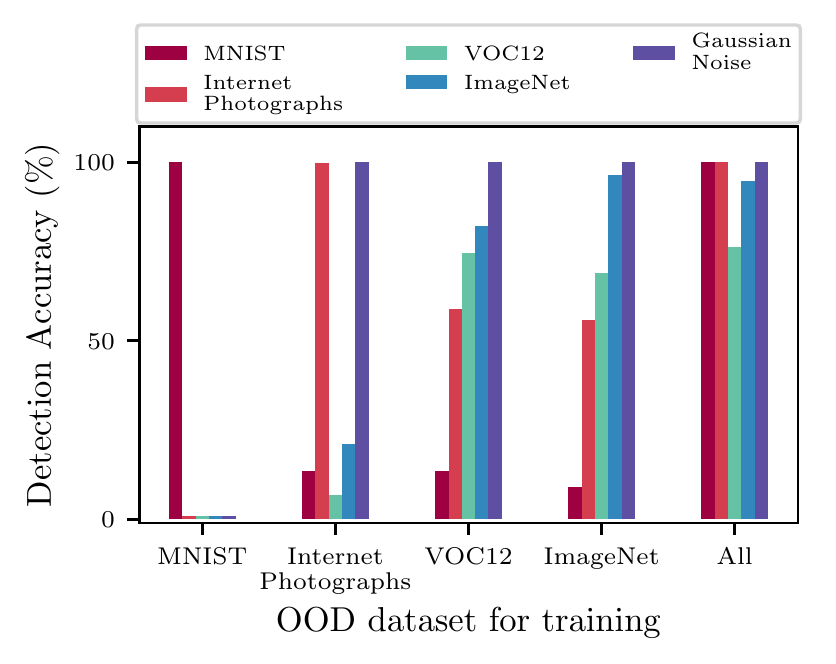}
    \caption{Classification accuracy of \emph{non-modified} OOD inputs (higher is better). The classification accuracy represents the percentage of OOD inputs  classified to the background class.}
   \label{fig:background_class_detection_scs}
\end{figure}

To train the classifier in presence of background class, we use 5,000 images from one of the datasets from MNIST, ImageNet, VOC12, and Random Photographs. The reason to include only 5,000 images is to avoid data bias when each class in the CIFAR-10 dataset has 5,000 images. To include multiple OOD datasets, we add one background class for each. Figure~\ref{fig:background_class_detection_scs} represent the success of the classifier in rejecting the \oodlong inputs. It shows that datasets such as VOC12 and ImageNet provide a high inter-dataset and intra-dataset generalization for detection of non-modified OOD inputs. Extensive experimental results for different attack methods, datasets are presented in Table~\ref{table:background_class}. 

\begin{table*}[t]
    \fontsize{15pt}{15pt}\selectfont
    \renewcommand{\arraystretch}{1.5}
    \caption{Adversarial training with a model augmented with additional background classes to discriminate between in and out distribution data. The target success rate is reported with \textsf{PGD-xent} attack using $\epsilon=8$ and random target label selection. We can see that  including a subset of an OOD dataset in training this model significantly decreases the target success rate of adversarial examples generated from this dataset. Surprisingly, this robustness can generalize to other OOD datasets, which are not included in the training. In addition, the robustness to multiple datasets can be achieved by including a subset of each in the training phase.}
    \label{table:background_class}
    \resizebox{\textwidth}{!}{%
        \begin{tabular}{ccccccccccccccccccccccccccccccccccccc}
            \toprule
            \multirow{3}{*}{\begin{tabular}[c]{@{}c@{}}OOD dataset \\ for training\end{tabular}} & \multirow{3}{*}{$\epsilon$}  && \multicolumn{4}{c}{MNIST} && \multicolumn{4}{c}{CIFAR-10} && \multicolumn{4}{c}{ImageNet} && \multicolumn{4}{c}{VOC12} && \multicolumn{4}{c}{\begin{tabular}[c]{@{}c@{}}Random \\ Photographs\end{tabular}} && \multicolumn{4}{c}{\begin{tabular}[c]{@{}c@{}}Gaussian \\ Noise\end{tabular}} \\ \cmidrule{4-7} \cmidrule{9-12} \cmidrule{14-17} \cmidrule{19-22} \cmidrule{24-27} \cmidrule{29-32} 
            &   && \multicolumn{2}{c}{PGD} & \multicolumn{2}{c}{CW} && \multicolumn{2}{c}{PGD} & \multicolumn{2}{c}{CW} && \multicolumn{2}{c}{PGD} & \multicolumn{2}{c}{CW} && \multicolumn{2}{c}{PGD} & \multicolumn{2}{c}{CW} && \multicolumn{2}{c}{PGD} & \multicolumn{2}{c}{CW} && \multicolumn{2}{c}{PGD} & \multicolumn{2}{c}{CW} \\  
            \cmidrule{4-7} \cmidrule{9-12} \cmidrule{14-17} \cmidrule{19-22} \cmidrule{24-27} \cmidrule{29-32} 
            &   && rand & ll & rand & ll && rand & ll & rand & ll && rand & ll & rand & ll && rand & ll & rand & ll && rand & ll & rand & ll && rand & ll & rand & ll\\ \midrule
            \multirow{3}{*}{None} & 1  && 1 & 0 & 1 & 0 && 1.3 & 0 & 1.3 & 0 && 3.2 & 0 & 2.6 & 0 && 3.5 & 0 & 3.5 & 0 && 3.6 & 0 & 4.5 & 0 && 11.0 & 0 & 8.4 & 0\\  
            & 8  && 5.1 & 0 & 5.7 & 0 && 22.9 & 5.1 & 22.3 & 4.4 && 44.9 & 4.9 & 39.5 & 4.6 && 54.9 & 11.5 & 50.3 & 9.9 && 46.3 & 12.2 & 47.5 & 11.4 && 88.6 & 6.1 & 87.8 & 4.6\\  
            & 16  && 11.9 & 0 & 12.5 & 0 && 65.6 & 48.6 & 60.4 & 42.7 && 83.8 & 53.9 & 80.3 & 46.9 && 91 & 74.2 & 87.8 & 66.9 && 83.9 & 63.7 & 82.1 & 57.1 && 100 & 100 & 100 & 100 \\ \midrule
            \multirow{3}{*}{MNIST} & 1 && 0 & 0 & 0 & 0 && 1.4 & 0 & 1.6 & 0 && 5.1 & 0 & 3.9 & 0 && 4.3 & 0 & 3.8 & 0 && 3.6 & 0 & 3.3 & 0 && 3.5 &  0 & 5.4 & 0\\  
            & 8  && 0 & 0 & 0 & 0 && 24.4 & 1.7 & 22.3 & 1.4 && 42.9 & 1.2 & 41.3 & 1 && 51.2 & 2.4 & 50.9 & 1.9 && 45.4 & 3.5 & 44.6 & 3.5 && 72.2 & 0 & 73.5 & 0 \\  
            & 16  && 0 & 0 & 0 & 0 && 65 & 28.5 & 61.6 & 24.8 && 82.9 & 17.7 & 81.7 & 14.6 && 90.4 & 33 & 89 & 27.3 && 82.3 & 38.3 & 80.3 & 31.6 && 100 & 0 & 100 & 0\\ \midrule
            \multirow{3}{*}{ImageNet} & 1 && 0.6 & 0 & 0.7 & 0 && 0.7 & 0 & 1 & 0 && 0 & 0 & 0.2 & 0 && 2.1 & 0 & 1.2 & 0 && 1.8 & 0 & 1.7 & 0 && 0 & 0 & 0 & 0\\  
            & 8 && 5.5 & 0 & 4.6 & 0 && 21.2 & 4.3 & 23.9 & 3.2 && 7.4 & 0.5 & 8.3 & 0.5 && 36.3 & 4 & 34.7 & 3 && 34.7 & 5.8 & 30.5 & 5 && 0 & 0 & 0 & 0\\ 
            & 16 && 13 & 0 & 13.4 & 0 && 63.7 & 47.3 & 60.3 & 40.7 && 39.1 & 13.6 & 36.7 & 12.3 && 84.8 & 56.6 & 80.2 & 48.3 && 77.2 & 52.5 & 72.2 & 44.3 && 5.4 & 0 & 6 & 0\\ \midrule
            \multirow{3}{*}{VOC12} & 1  && 0.5 & 0 & 0.4 & 0 && 0.5 & 0 & 1.4 & 0 && 0.6 & 0 & 1 & 0 && 1.1 & 0 & 1.3 & 0 && 1.8 & 0 & 2.6 & 0 && 0 & 0 & 0 & 0\\  
            & 8  && 5.1 & 0 & 5.2 & 0 && 23.6 & 3.7 & 21.9 & 3.7 && 15.8 & 1.2 & 17.8 & 3.7 && 32.2 & 2.6 & 31.8 & 2.6 && 33.1 & 7 & 34.4 & 5.3 && 24 & 0 & 24.7 & 0\\  
            & 16  && 12.4 & 0 & 13.9 & 0 && 65.5 & 46.4 & 64.8 & 46.7 && 54.5 & 22 & 56.8 & 21.9 && 80.7 & 52 & 80.3 & 52 && 75.5 & 49.5 & 79.9 & 51.5 && 100 & 100 & 100 & 100\\ \midrule
            \multirow{3}{*}{\begin{tabular}[c]{@{}c@{}}Random \\ Photographs\end{tabular}} & 1  && 1 & 0 & 0.7 & 0 && 1 & 0 & 1.5 & 0 && 2.5 & 0 & 2.9 & 0 && 3.2 & 0 & 2.4 & 0 && 0 & 0 & 0 & 0 && 0 & 0 & 0 & 0\\  
            & 8  && 6.6 & 0 & 5.8 & 0 && 23.2 & 3.3 & 20.9 & 2.4 && 39.1 & 4 & 36.8 & 3.3 && 54.5 & 8.7 & 51.1 & 7.8 && 0.9 & 0.4 & 0.7 & 0.4 && 42.3 & 24 & 45.5 & 16.4 \\  
            & 16  && 16.4 & 0 & 13.8 & 0 && 63.8 & 42.6 & 61.1 & 36 && 81 & 49.6 & 76.6 & 41.6 && 90.8 & 69.9 & 87.1 & 62.5 && 38.5 & 30.2 & 36.6 & 29.3 && 100 & 100 & 100 & 100\\ \midrule
            \multirow{3}{*}{\begin{tabular}[c]{@{}c@{}} All \end{tabular}} & 1  && 0 & 0 & 0.7 & 0 && 1.1 & 0 & 1.4 & 0 && 0.2 & 0 & 0.2 & 0 && 1 & 0 & 0.8 & 0 && 0 & 0 & 0 & 0 && 0 & 0 & 0 & 0\\  
            & 8  && 0 & 0 & 0 & 0 && 21.1 & 1.2 & 20.2 & 1 && 8.6 & 0.2 & 10.5 & 0.2 && 31.6 & 1 & 31 & 1 && 1.3 & 0.6 & 1.4 & 0.6 && 0 & 0 & 0 & 0 \\  
            & 16  && 0 & 0 & 0 & 0 && 65.6 & 24.7 & 63.7 & 19 && 49 & 9.6 & 45.9 & 7.8 && 80.2 & 28.3 & 77.1 & 23.5 && 43.7 & 11.4 & 41.3 & 10.1 && 51
            4& 2.1 & 46.9 & 0.2\\ \bottomrule
        \end{tabular}
    }
\end{table*}

\end{document}